\documentclass[10pt,twocolumn,letterpaper]{article}

\usepackage{cvpr}
\usepackage{times}
\usepackage{epsfig}
\usepackage{graphicx}
\usepackage{amsmath}
\usepackage{amssymb}
\usepackage{multirow}
\usepackage{subfigure}
\usepackage{type1cm}
\usepackage{color}
\usepackage{ulem}
\usepackage{booktabs}
\usepackage{authblk}

\usepackage[pagebackref=true,breaklinks=true,letterpaper=true,colorlinks,bookmarks=false]{hyperref}

\cvprfinalcopy 

\def\cvprPaperID{3605} 

\ifcvprfinal\fi
\begin{document}

\title{Defocus Blur Detection via Salient Region Detection Prior}

\author[1]{Ming Qian}
\author[1]{Min Xia}
\author[2]{Chunyi Sun}
\author[1]{Zhiwei Wang}
\author[1]{Liguo Weng}

\affil[1]{Nanjing University of Information Science \& Technology, China \authorcr  \{mingqian,xiamin,wangzhiwei,lweng\}@nuist.edu.cn}
\affil[2]{Australian National University, Australia 
\authorcr  u6342253@anu.edu.au}




\maketitle

\begin{abstract}
Defocus blur always occurred in photos when people take photos by Digital Single Lens Reflex Camera(DSLR),  giving salient region and aesthetic pleasure.
Defocus blur Detection aims to separate the out-of-focus and depth-of-field areas in photos, which is an important work in computer vision. 
Current works for defocus blur detection mainly focus on the designing of networks, the optimizing of the loss function, and the application of multi-stream strategy, meanwhile, these works do not pay attention to the shortage of training data.
In this work,  to address the above data-shortage problem, we turn to rethink the relationship between two tasks: defocus blur detection and salient region detection.
In an image with bokeh effect, it is obvious that the salient region and the depth-of-field area overlap in most cases. So we first train our network on the salient region detection tasks, then transfer the pre-trained model to the defocus blur detection tasks. 
Besides, we propose a novel network for defocus blur detection. Experiments show that our transfer strategy works well on many current models, and demonstrate the superiority of our network.
\end{abstract}

\section{Introduction}
\par Defocus blur detection(DBD) aims to separate the out-of-focus and the depth-of-field areas in photos, these photos with bokeh effects are usually taken by DSLR with a wide aperture lens\cite{davis2008practical}.

\par DBD is an essential work for many computer vision tasks, as it has latent relationships with several tasks, such as salient region detection\cite{qin2020u2}, rendering realistic bokeh\cite{qian2020bggan}, quality assessment\cite{fang2020perceptual}, image deblurring\cite{li2020efficient}, and so on.

\par During recent years, there are various DBD methods have been proposed. It can be divided into two categories: hand-craft based methods and deep learning-based methods, we will discuss these methods in Related Work with more details.

\begin{figure}[!ht]
\begin{center}
\includegraphics[width=0.9\linewidth]{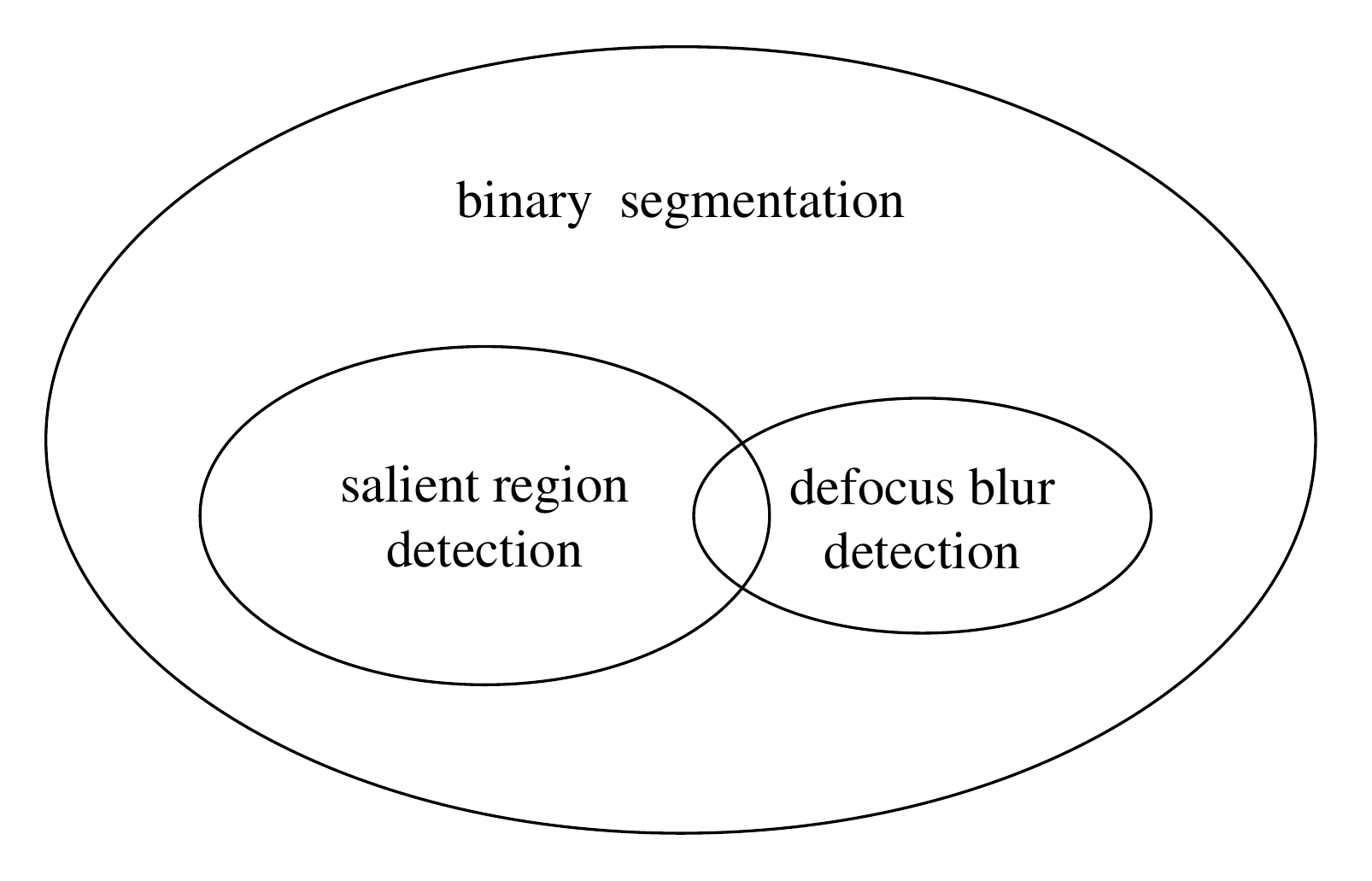}
\end{center}
   \caption{Domain relationship between SOD \& DBD}
\label{fig:domain}
\end{figure}

\par Current deep learning based methods are more focused on the design of model structures and loss functions. Those methods ignore the fact that the number of training samples in DBD dataset is quite limited. For instance, the training dataset used by current deep learning-based methods is Shi et al.'s dataset\cite{shi2014discriminative}, which contains 704 defocus blur images, previous works often split them into two categories, images for training and for testing, 604 and 100 separately. Compared with other computer vision tasks which usually contain thousands even billions of training samples, the number of DBD training samples is not sufficient.

\begin{figure*}[!ht]
\begin{center}
\includegraphics[width=0.8\linewidth]{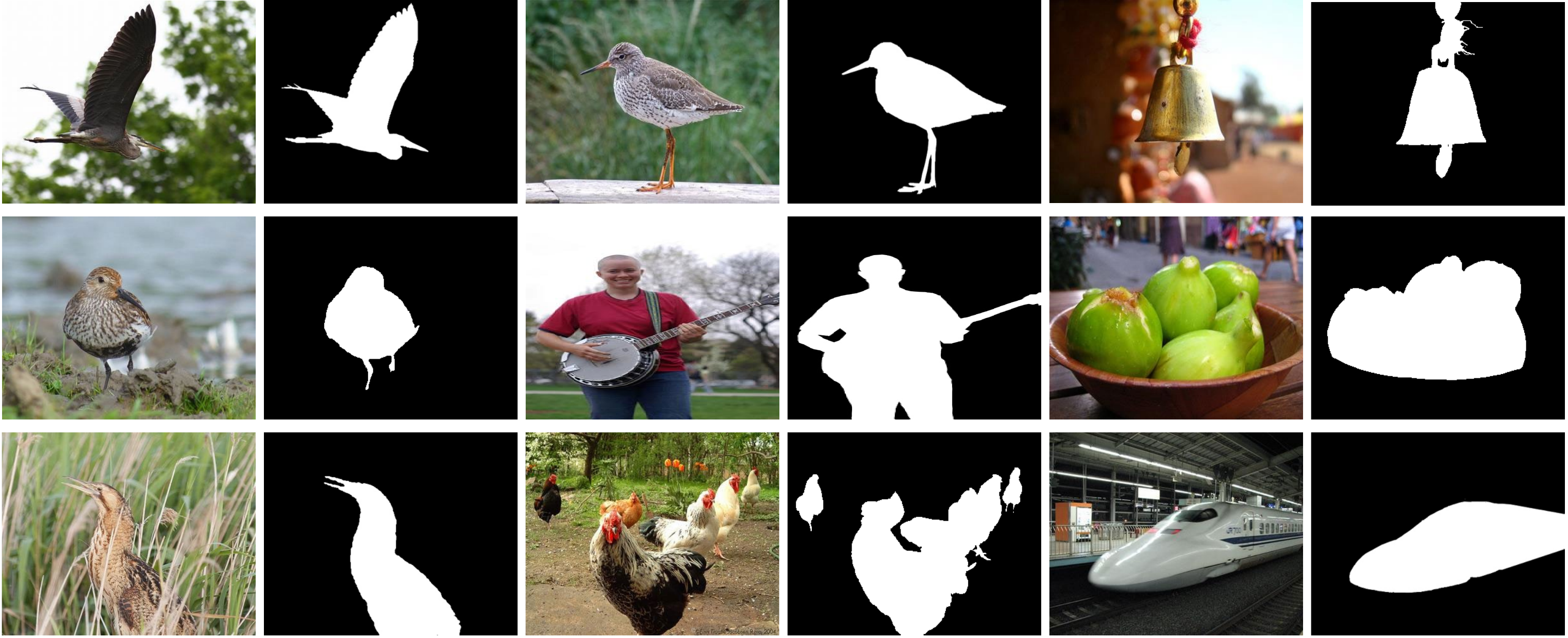}
\end{center}
   \caption{Example of DUTS-TR dataset for SOD}
\label{fig:SOD_example}
\end{figure*}

\begin{figure*}[!ht]
\begin{center}
\includegraphics[width=0.8\linewidth]{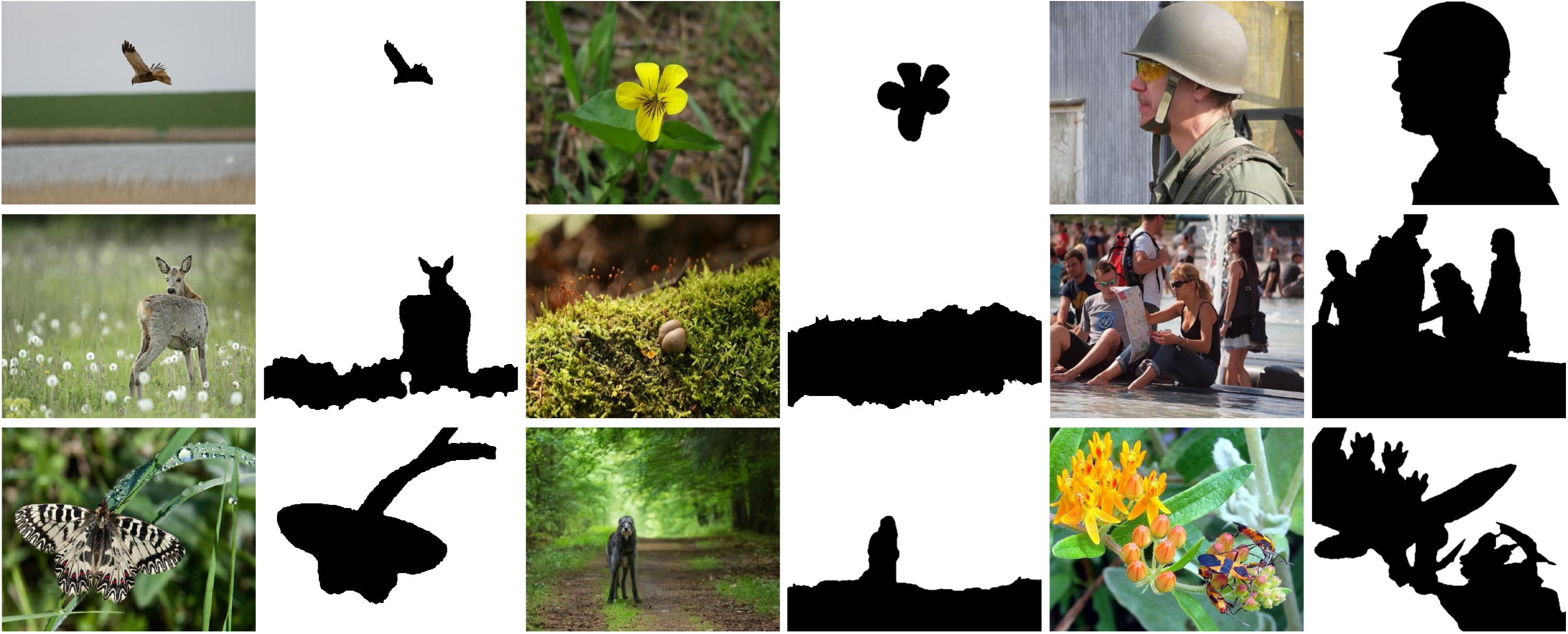}
\end{center}
   \caption{Example of Shi et al.'s dataset for DBD}
\label{fig:DBD_example}
\end{figure*}

\par Salient region detection(SOD) is another prosperous region of computer vision, it aims at detecting and segmenting the most salient objects in the image. Coincidentally, there are some similarities between these two areas. Firstly, SOD and DBD are both binary segmentation. In other words, almost all deep learning models designed for SOD or DBD task can easily be applied to each other via retraining. From Figure.\ref{fig:SOD_example}, we can find that 255 represents salient regions $s$ and 0 represent background regions on salient region masks $g$. Meanwhile, from Figure.\ref{fig:DBD_example}, 255 represents defocus blur regions $d$ and 0 represents non-defocus regions on defocus region masks. Besides, non-defocus regions are also named bokeh regions $b$. Figure.\ref{fig:SOD_example} and Figure.\ref{fig:DBD_example} are choosed from Shi et al.'s dataset\cite{shi2014discriminative} and DUTS-TR\cite{wang2017learning}. 
Secondly, those two tasks are very similar or even the same in some scenarios. In the image group in the upper right corner of Figure.\ref{fig:DBD_example}, the black area, the soldier, namely, the 0-pixel region, is obviously a salient region in the SOD subject.  But the bokeh region in the first image group in the second row is not identical to its salient region, because some grasses and flowers are in the bokeh region, but the salient region in SOD tasks should not contain these grasses and flowers. Besides, this phenomenon widely exist in the other images in the two datasets.

\par We can draw a conclusion from this observed phenomenon: in photos with bokeh effects taken by DSLR, the bokeh region and the salient region are overlapping, and sometimes they are identical, which is represented as:
\begin{equation}
s \cap d \neq \o
\end{equation}
\par In Figure.\ref{fig:domain}, we describe the relationship between binary segmentation, SOD and DBD. It can be seen that SOD and DBD are subtasks of binary segmentation, and SOD tasks and DBD tasks are closely related. 

\par Current development of the SOD task is faster than that of the DBD task. On the one hand, the DUT-TR dataset is the common training dataset for most SOTA models in SOD, which contains about 10553 images.  It is far more than Shi et al.'s dataset. Plus, there are more elegant and fancy ideas in current SOD jobs due to the prosperity of the SOD task, there are more researchers who are interested in SOD task and more fancy models proposed for SOD task.

\par In this work, we hypothesis that leverage the use of SOD prior could alleviate the data shortage of the DBD task and take advantage of the well-developed network structure of SOD task. Furthermore, we draw on the experience of elegant SOD models and design a novel model to capture the deeper information of different levels for DBD task. Experiments show that our strategy quantitatively and qualitatively outperforms many previous works. In addition, due to the highly correlated prior provided by salient region detection, we can make the DBD model to converge faster.

\par Overall, our main contributions can be summarized as follows:
\begin{itemize}
\item We find an elegant way to improve the DBD results: training our model on SOD dataset first, and then utilize the pre-trained model as the prior to fine-tuning the model on DBD dataset, which can markedly improve the final results and reduce the training time.
\item We propose a novel network architecture named Adaptive Feature Integration UNet(AFIU)  to solve the scale variance problem in the salient object detection task. This architecture achieves great performance on SOD and DBD tasks.
\item We will release our results and code publicly to promote the development of SOD\&DBD tasks.
\end{itemize}

\section{Related Work}
\subsection{Defocus Blur Detection}
Defocus blur Detection aims to separate the out-of-focus and depth-of-field area in photos, which is a fundamental work in computer vision. Thus, various models have been proposed to solve these problems. In early works, they mainly focus on the extracting of hand-crafted features\cite{su2011blurred,shi2014discriminative,pang2015classifying,shi2015just,yi2016lbp,tang2016lspectral,golestaneh2017spatially,xu2017estimating}. With the development of deep learning, CNN based methods have upset the ecological balance of many computer vision tasks, such as DBD task. DHDE was the first work extracting CNN based features on DBD task\cite{park2017unified}, it extracted hand-crafted features and CNN based features, and combined them together into a fully convolutional classifier to detect the blur region. Purohit et al.\cite{purohit2018learning}
proposed a double-branch structure to extract global features and local features separately. Zhang et al.\cite{zhang2018learning} proposed a dilated fully convolutional neural network, which contains a pyramid pooling module and edge refinement module to improve the capability of the model. 
BTBNet\cite{zhao2018defocus} proposed a multi-stream bottom-top-bottom fully convolutional network, which includes two streams, a low-level stream and a high-level semantic information stream separately. DeFusionNET\cite{tang2019defusionnet} detected blur regions via recurrently fusing and refining multi-scale deep features, which can effectively extract multi-scale features. CENet\cite{zhao2019enhancing} divided DBD detector into multitudinous smaller detectors and proposed a cross-ensemble network to reduce the detection error of small classifiers. R$^2$MRF\cite{tang2020r2mrf} improves the capability of extracting multi-scale information of the model via recurrently refining multi-Scale residual features.

\subsection{Salient Region Detection}
\par People could detect visually distinctive, namely, salient.  Scene regions effortlessly and repeatedly in a pre-attentive stage\cite{borji2019salient}.
Classical approaches for salient object detection are based on hand-crafted features \cite{Hu2005RobustSA,Li2017CNNFS,Achanta2012SLICSC,Lee2016DeepSW}, and use multi-layer perception to classify these features like pixels and small sub-regions. Those traditional methods are not robust to scale changes and fail in complex scenes easily. Fully Convolutional Networks (FCN) gain great attention to many computer vision tasks and become a dominating approach in the salient object detection task. A good result of salient map detection depends on how effective multi-level features are aggregated. Previous works have provided manifold approaches to aggregate multi-level features. Our basic skeleton U-Net fuses multi-level features directly via transpose convolution and concatenation. Direct concatenation is simple, but it may lack the ability to aggregate the most important information from previous layers. It may also fuse redundant information as those features come from the same scale level.  LEGS \cite{Wang2015DeepNF} proposes a network architecture that combines local estimation and global proposals via two sub-networks and introduces an extra global search module to fuse the estimation from those two sub-networks. This approach enlarges the network size significantly. Also, performing local and global predictions separately may not aggregate the multi-level features extracted by adjacent layers efficiently. \cite{Zhang2018ABM} uses a bi-directional message-passing model to integrate multi-level information and use a gate function to control message passing; \cite{Zhang2018ProgressiveAG} leverage the attention module to select the most informative feature from multi-level architecture by giving it large weights. Compared to those two approaches, our network allows gaining the feature map in one-pass while still fusing multi-level features very well.

\subsection{Deep Transfer Learning}
\par Deep Transfer Learning can be classified into four categories\cite{tan2018survey}: instance-based deep transfer learning\cite{xu2017unified}, mapping-based deep transfer learning\cite{long2015learning}, network-based transfer learning\cite{oquab2014learning}, and adversarial-based deep transfer learning\cite{tzeng2017adversarial}. We mainly focus on the network-based transfer learning method.
\par Network-based transfer learning method means reusing part of network weights on source domain, including connecting weights and model skeleton, and then transfer to a part of networks on the target domain. Normally, the reusing part of pre-trained models is the front-layers, which can be treated as a feature extractor. \cite{huang2013cross} divides the network into two parts. The first part is language-independent, and the last layer is a language-relative classifier. The former can be transformed into various languages. \cite{oquab2014learning} uses the front-layers of CNN trained on ImageNet dataset\cite{deng2009imagenet} to calculate mid-level image representations on other datasets. 
Various works underline the benefit can be taken from fine-tune strategy from pre-trained models trained on big datasets\cite{kading2016fine}, such as image retrieval\cite{yan2020deep}, semantic segmentation\cite{ding2020semantic}, object detection\cite{redmon2018yolov3}, salient region detection\cite{MINet-CVPR2020} and defocus blur detection\cite{tang2019defusionnet}. While most current DBD works used well-trained models trained on ImageNet as the backbone and achieved a positive effect, we think that compared with image classification, SOD has a stronger correlation with DBD as mentioned above. So in this work, we select SOD model as a pre-trained model to fine-tune our model.

\section{Proposed Method}
\subsection{Transfer Method}
\par Previous works only train the model on DBD datasets from scratch, which ignores the fact that the figure of the DBD dataset training samples is limited. 
To get the best performance of a model $f$, we select the pre-training method as our transfer method, as it is the commonly used, convenient and sufficient approach\cite{he2019rethinking}. We first train a model $f$ on DUT-TR\cite{wang2017learning}, then the well-trained model $f$  is used to initialize the parameters of the whole network. 
We train the above model on Shi et al's dataset\cite{shi2014discriminative}. Similar to ImageNet pre-training, before the model trained on the DBD training set, we firstly train the model on the SOD training set.

\subsection{Model}
\begin{figure}[h]
\centering
\includegraphics[width=0.5\textwidth]{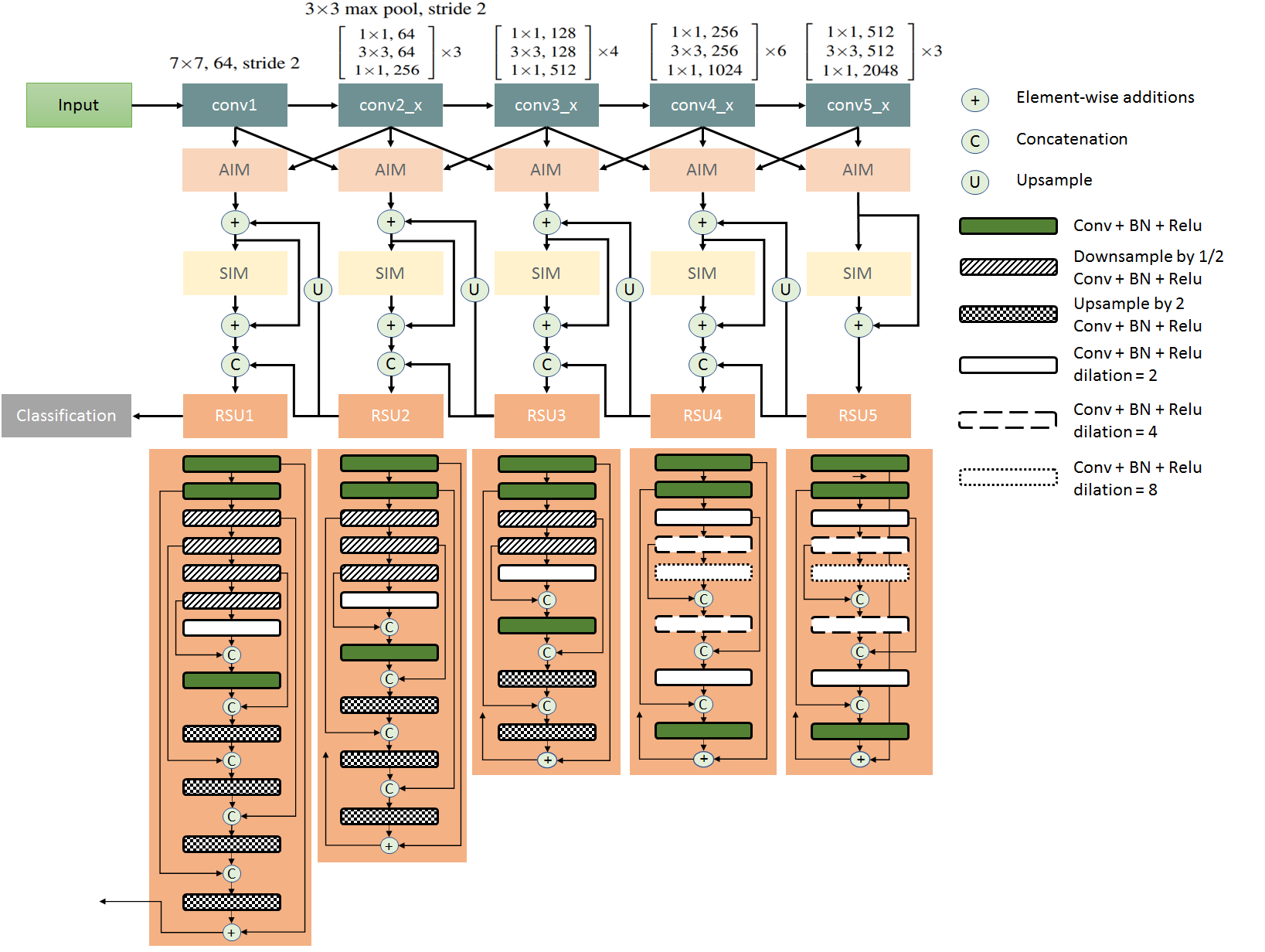}
\caption{Model Architecture}
\label{overall_architecture}
\end{figure}

\par The overall architecture of our network is shown in Figure.\ref{overall_architecture}. The encoder is formed by blocks consisting of an AIM, a SIM, and an extra fusion module at the end. The decoder is formed by a stacked U-net with residual connections. In this section, we will explain our network architecture in a top-down manner, we will first explain the usage of each module (encoder, decoder, etc).
\par Our network is built upon the ResNet-50 \cite{He2016DeepRL} backbone. In the following demonstration, we use $E_x$ to refer to the corresponding levels in ResNet-50. Eg. $E_1$ represents $conv1$ in ResNet-50, $E_2$ represents $conv2\_x$ in ResNet-50, etc. Multi-level features will first be sent to AIM modules in the encoder. AIM module is designed to provide relevant and effective supplement by combining features from adjacent levels. Then, the outputs from AIMs are fed individually into separate SIM modules for extracting multi-scale information from the current level adaptively. Finally, the processed information will be fused via a convolutional module and sent into the decoder. The decoder is designed as a stacked U-net to capture intra-stage multi-scale features.
\par The main components in the encoder module include aggregate interaction modules (AIMs) and self-interaction modules (SIMs) as proposed in \cite{Pang2020MultiScaleIN}.After receiving different resolutions of features learned by different levels in Resnet50, we apply AIMs for aggregating features from these different levels, and then apply SIMs for further feature extraction. Each AIM module receives 2 or 3 different levels features from Resnet50 and the output from the AIM module will become the input into the SIM module.  For the SIM module, it first applies convolutions of different parameters to the input to obtain two branches with different feature resolutions. It then performs up-sampling to the branch with low resolution and adds it to the original high-resolution branch via element-wise addition. Meanwhile, it also performs down-sampling to the branch with high resolution and adds it to the original low-resolution branch via element-wise addition.
\par Such interaction between features of different resolutions can help obtain abundant information about variant scales. The objective of SIM module is for acquiring multi-scale features from integrated features.
In Figure.\ref{overall_architecture}, it includes five U-Net-like decoders. In the deepest level (the level corresponds to $E_5$ (conv5\_x)), the feature output(${E_5}^*$) from ResNet-50 will first be sent into the encoder, and the output ($EC_5$) from the encoder will be sent to RSU5 block, which outputs decoded feature($D_5$). In the intermediate levels, similarly, before sent to the U-Net block, the feature (${E_x}^*$) extracted from the backbone will first be feed into our adaptive feature fuse encoder. Then, we combine this encoded feature ($EC_x$) with the decoded feature ($D_{x+1}$) from the deeper level by concatenation. This combined feature will be feed into the UNet block at the current level for decoding.  In the most shallow level (the level corresponds to $E_1$ (conv1)), after getting the decoded feature from RSU2, the final output will be obtained via a ${1}\times {1}$ convolution layer.
\par The details of each RSU module are illustrated on the bottom of Figure. \ref{overall_architecture}, the legend on the left indicates the elements with the corresponding sub-module.

\begin{figure*}[!ht]
	\centering
    \begin{minipage}[t]{0.24\textwidth}
    \centering
    \includegraphics[width=4.5cm]{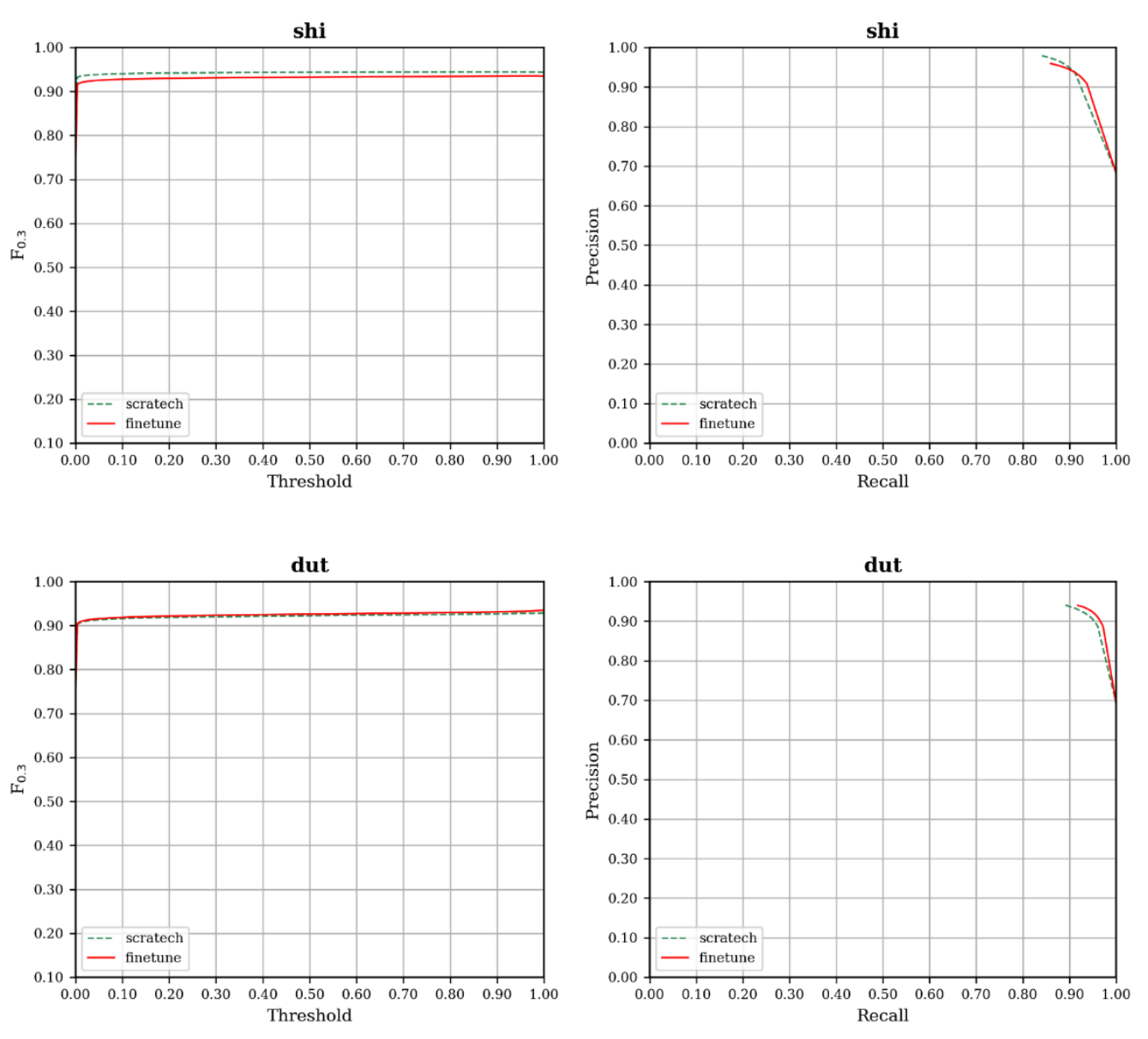}
    \centerline{(a) BAS}
    \end{minipage}
    \begin{minipage}[t]{0.24\textwidth}
    \centering
    \includegraphics[width=4.5cm]{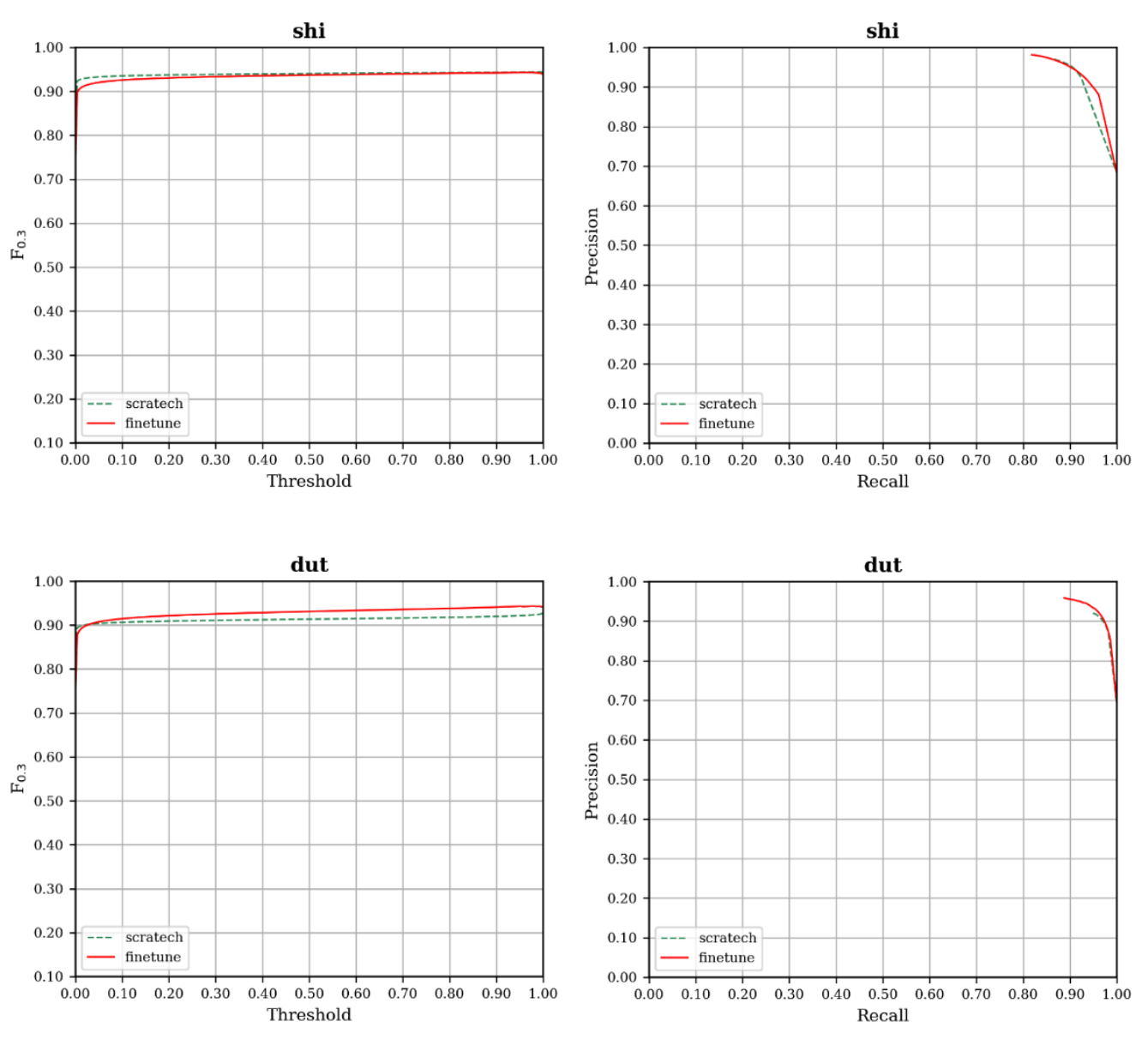}
    \centerline{(b) CPD}
    \end{minipage}
    \begin{minipage}[t]{0.24\textwidth}
    \centering
    \includegraphics[width=4.5cm]{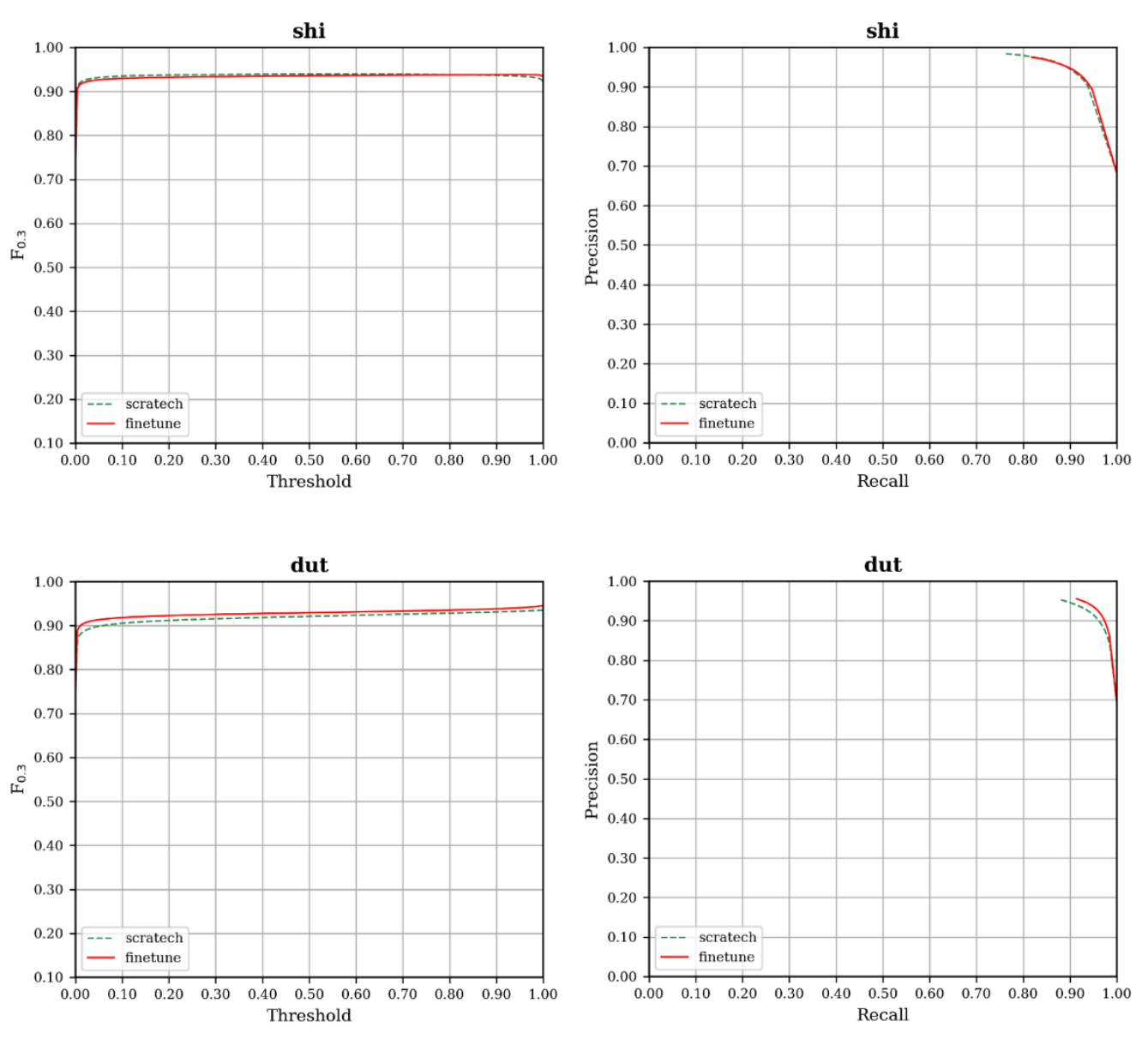}
    \centerline{(c) MIN}
    \end{minipage}
    \begin{minipage}[t]{0.24\textwidth}
    \centering
    \includegraphics[width=4.5cm]{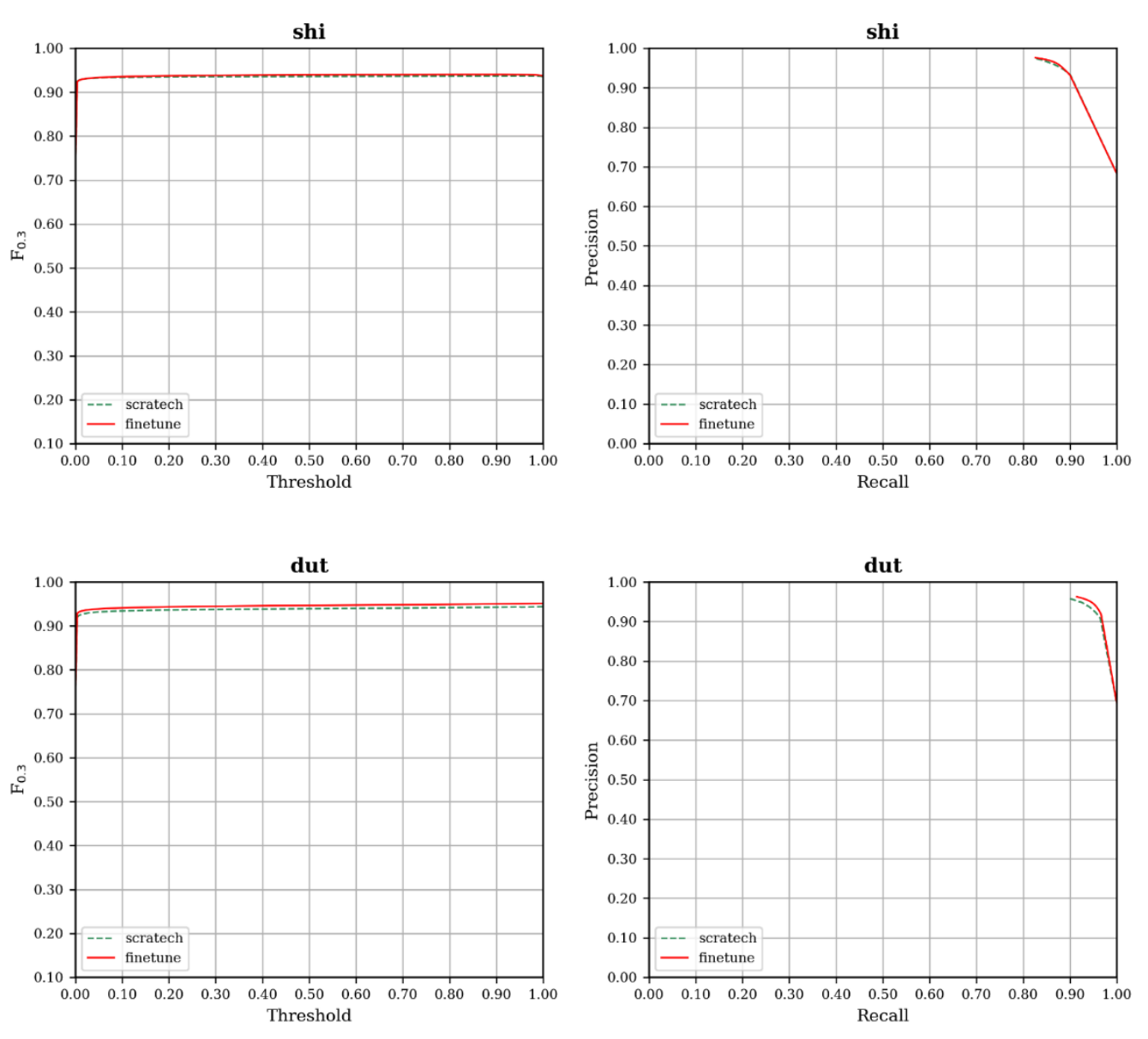}
    \centerline{(d) OURS}
    \end{minipage}

	\centering
	\caption{Illustration of PR curves and the F measure curves on the two datasets of some methods with different strategy.}
	\label{fig:PRFM}
\end{figure*}

\subsection{Implementation Details}
\par To get a well-trained model, we first train the model on the SOD training set from scratch, and then finetune the above model on DBD dataset.
\par When train our model, we select DUTS-TR \cite{wang2017learning} as our SOD training dataset. The images are resized to $320*320$ with 50\% probability of vertical flip. Also, Random Color jittering is used for data augmentation. We choose Adam as our optimizer, with the initial learning rate set to $1e^{-5}$, the betas set to (0.9,0.999), and the weight decay set to 0. We set the batch size as 8. The network is converged after around 150 epochs(24 hours) and 200 epochs(6 hours) in the first stage and finetune stage separately. The loss function we use is standard binary cross-entropy,
\begin{equation}
\centering
\begin{split}
L = - \sum_{(r,c)}^{(h,w)} [P_{G(r,c)}\log{P_{S(r,c)}} + (1- P_{G(r,c)}) \\
\log{(1-P_{S(r,c)})}]
\end{split}
\end{equation}
where $(r,c)$ indicates the coordinates in the salient maps and $(h,w)$ indicates the output size; $P_G$, $P_S$ denote for the ground truth salient map and predicted salient map, respectively.
\par In the fine-tuning stage, we train the model just follow the details of training our proposed model on SOD training set. the only differences are the figure of training epoch which is about 200 and we need to load the model weights trained on SOD dataset as initial weights rather than train from scratch.
\par Our training and testing are calculating on a 6-core, 12 threads PC with an INTEL i7-8700 3.2 GHz CPU (64GB RAM) and a GTX 2080ti GPU (11GB memory). We will release our code and results maps later.

\section{Experiments}
\subsection{Datasets}
\subsubsection{SOD datasets}

\par For the SOD pre-training, our network is trained on the DUTS-TR \cite{wang2017learning} dataset which contains 10533 images. It is a widely-used large-scale training dataset for salient object detection tasks. We choose to evaluate our method on six commonly used datasets, DUT-OMRON \cite{yang2013saliency}, DUTS-TE \cite{wang2017learning}, ECSSD \cite{Shi2016HierarchicalIS}, HKU-IS \cite{Li2015VisualSB}, PASCAL-S \cite{Li2014TheSO} and SOD \cite{Li2018ABD}, to make comparison with other state-of-the-art methods.
\par DUT-OMRON, DUT-TE, ECSSD and HKU-IS are large-scale datasets with 5168, 5019, 1000 and 4445 test images, respectively. DUT-OMRON and DUT-TE contain rich object subjects with many complex scenes. Images in HKU-IS dataset have larger diversity on the spatial distribution and many of these images have indistinguishable backgrounds; ECSSD contains massive images taken in complex environments, therefore with high-level semantic meaning; PASCAL-S contains 850 challenging images with multiple objects and background clutter; SOD only contains 300 images but it is the most challenging one, where object overlapping happens frequently. Thus, we include all of these datasets to test the generalization capacity and accuracy of our method from different aspects.
\subsubsection{DBD datasets}
\par We train our model on Shi et al's dataset\cite{shi2014discriminative}, Shi et al's dataset contain 704 images with defocus blur and corresponding masks. We select 604 for training and the rest for evaluation. Besides, We use DUT\cite{zhao2018defocus} for evaluation. DUT contains 500 images with pixel-wise annotations. This dataset is more challenging because some images may have low contrast focal regions, homogeneous regions, and background clutter.

\begin{figure*}[ht]
	\centering
    \begin{minipage}[t]{0.48\textwidth}
    \centering
    \includegraphics[width=6cm]{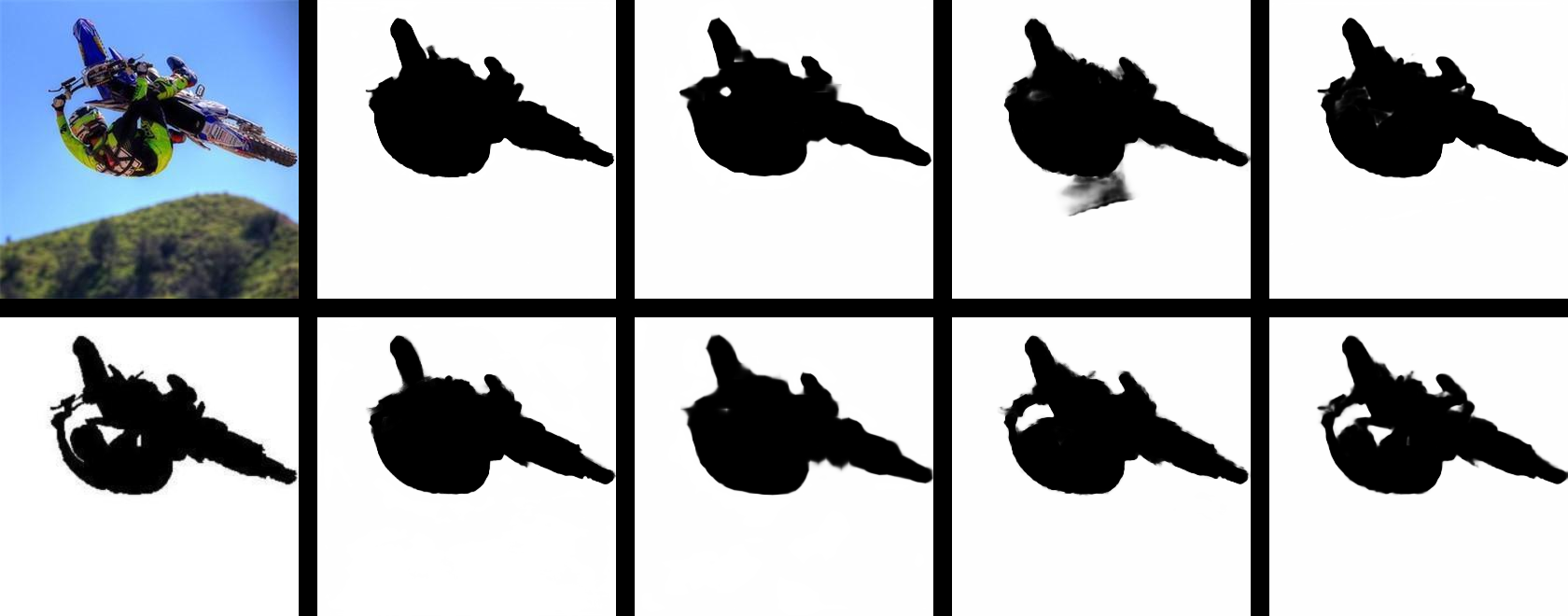}
    \end{minipage}\hspace{-15mm}
    \begin{minipage}[t]{0.48\textwidth}
    \centering
    \includegraphics[width=6cm]{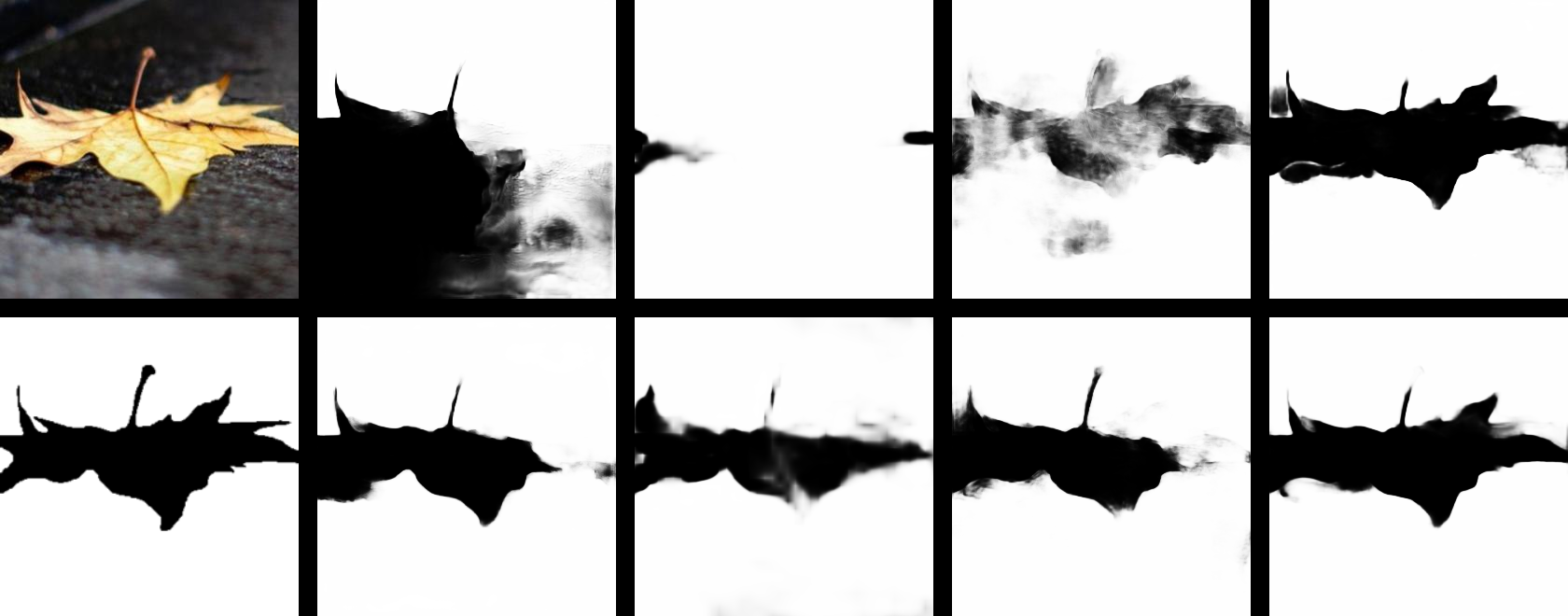}
    \end{minipage}  \vspace{0.5mm}
    
    \begin{minipage}[t]{0.48\textwidth}
    \centering
    \includegraphics[width=6cm]{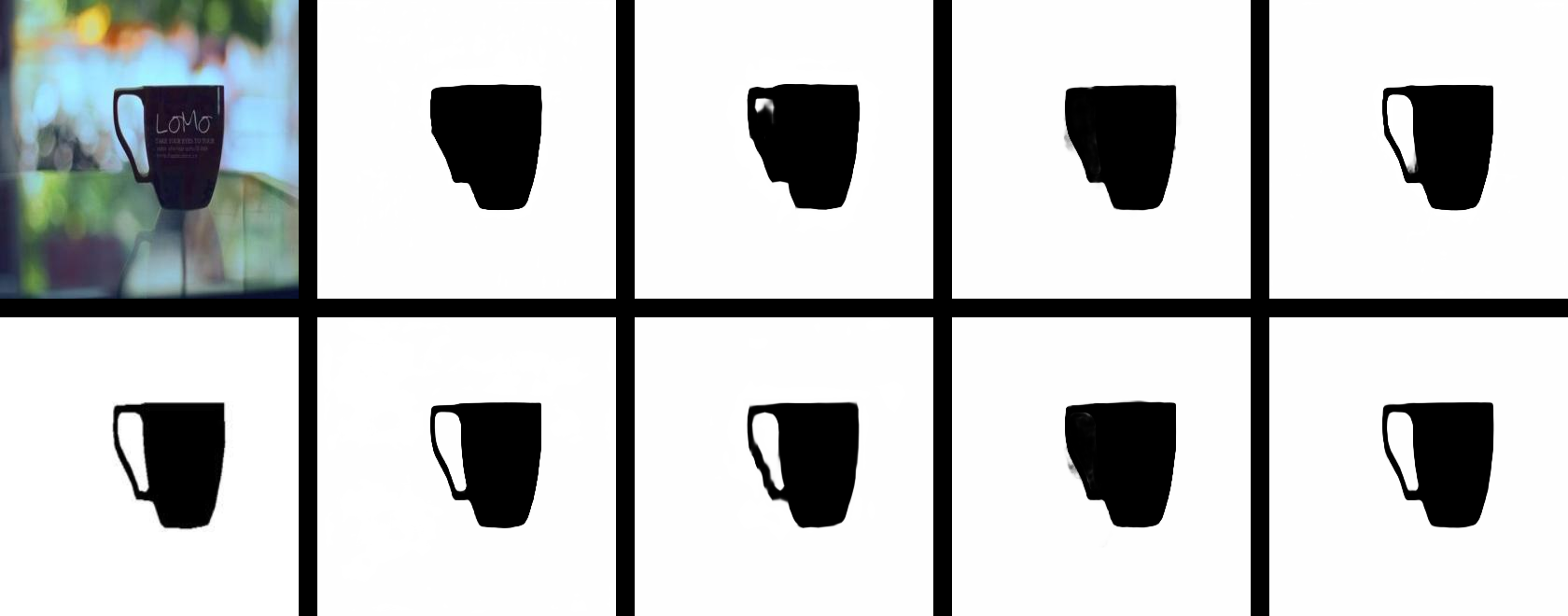}
    \end{minipage}\hspace{-15mm}
    \begin{minipage}[t]{0.48\textwidth}
    \centering
    \includegraphics[width=6cm]{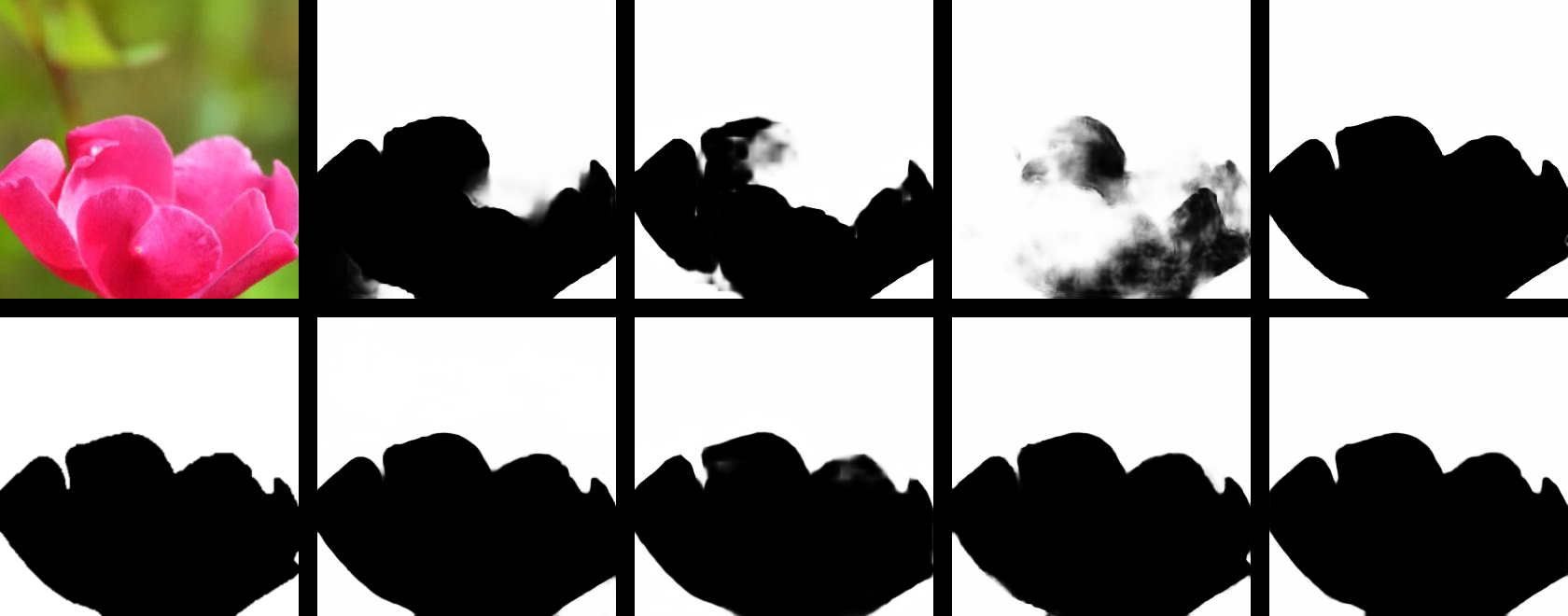}
    \end{minipage}  \vspace{0.5mm}
    
    \begin{minipage}[t]{0.48\textwidth}
    \centering
    \includegraphics[width=6cm]{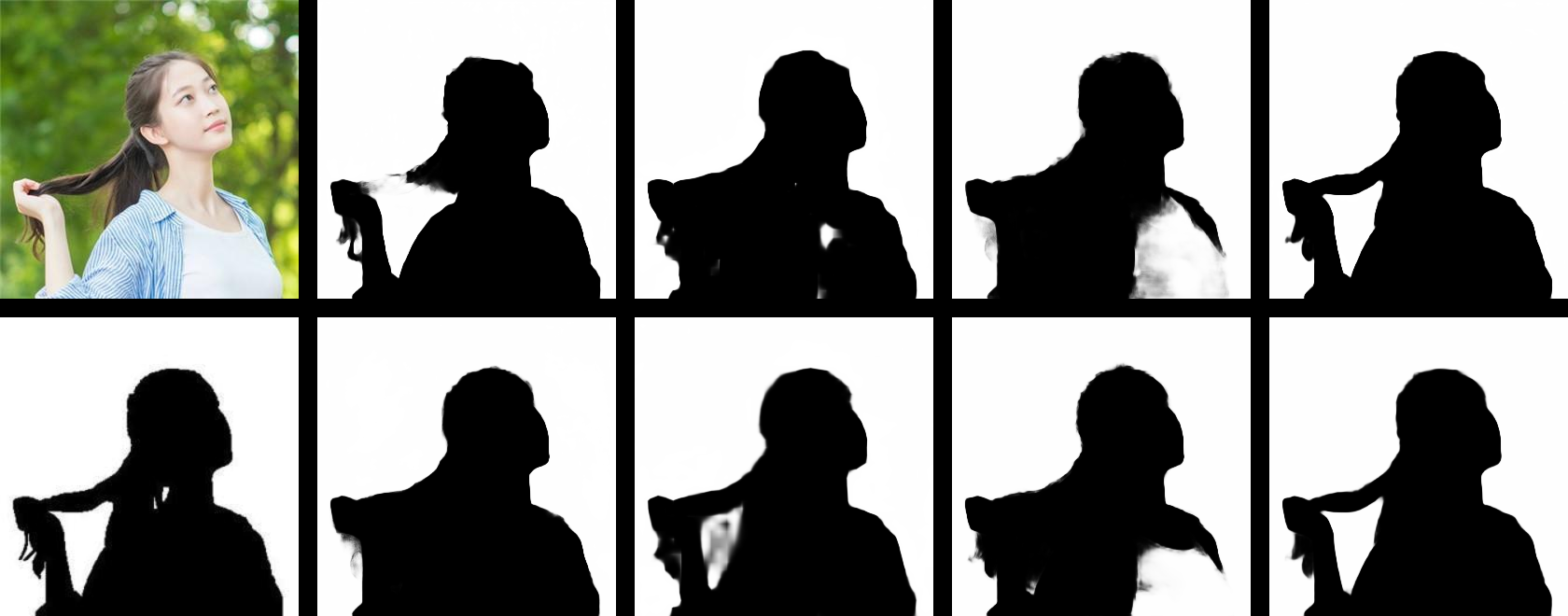}
    \end{minipage}\hspace{-15mm}
    \begin{minipage}[t]{0.48\textwidth}
    \centering
    \includegraphics[width=6cm]{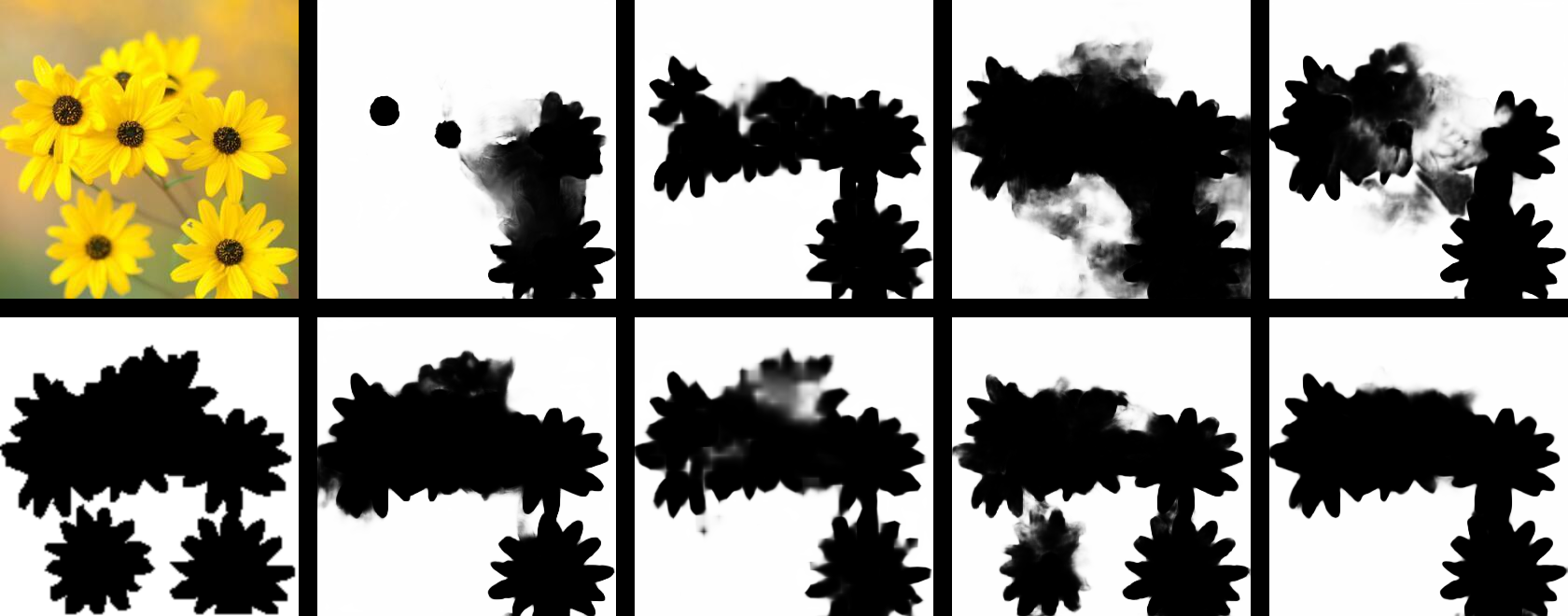}
    \end{minipage}  \vspace{0.5mm}
    
    \begin{minipage}[t]{0.48\textwidth}
    \centering
    \includegraphics[width=6cm]{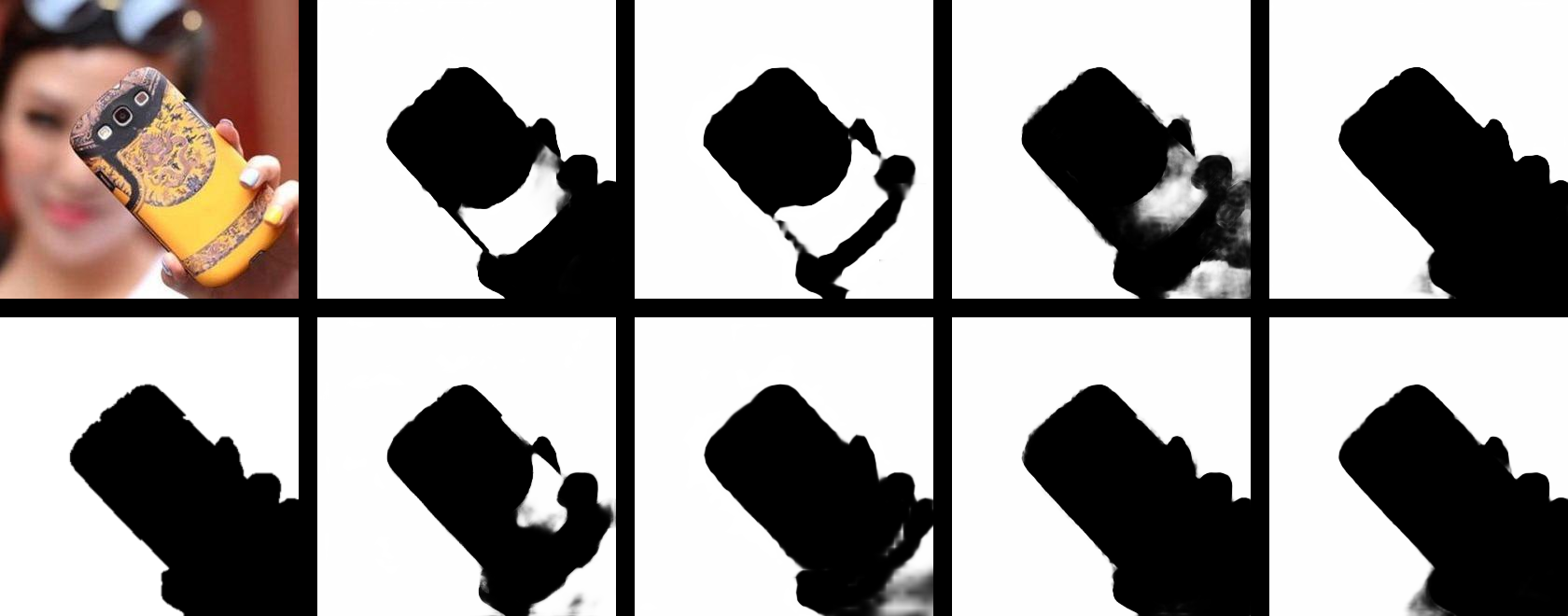}
    \end{minipage}\hspace{-15mm}
    \begin{minipage}[t]{0.48\textwidth}
    \centering
    \includegraphics[width=6cm]{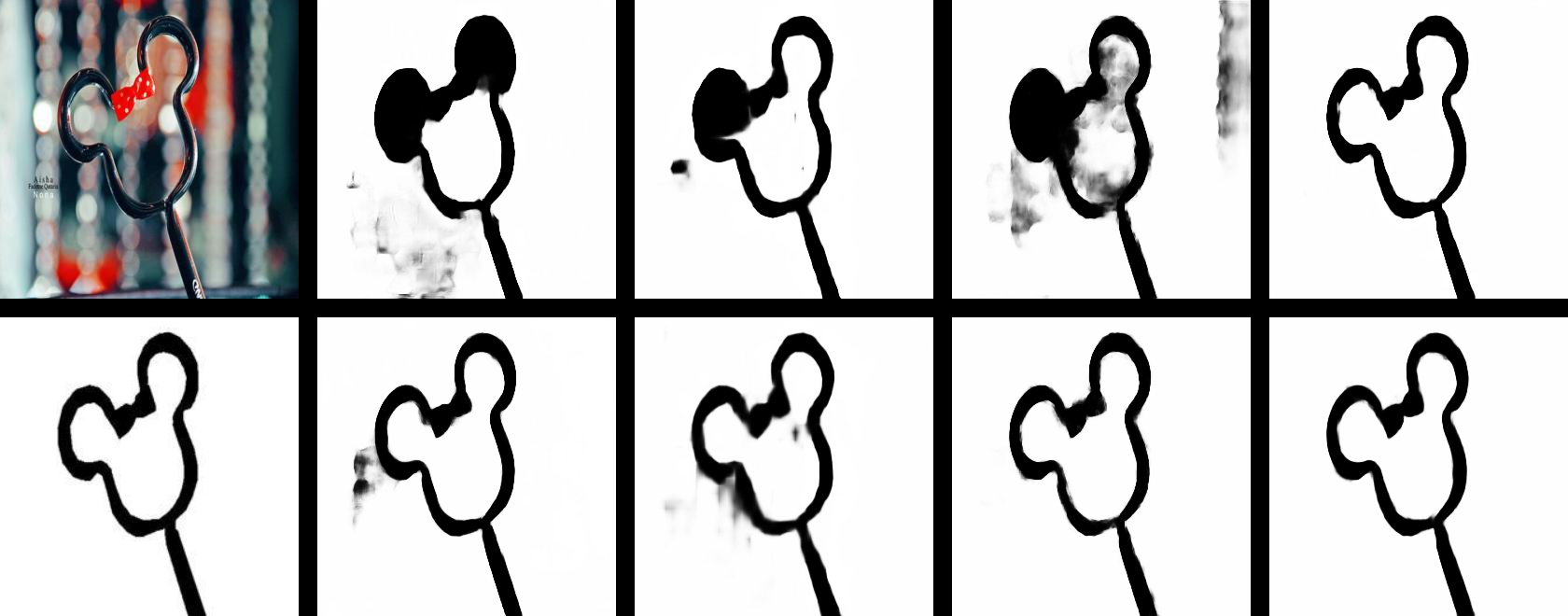}
    \end{minipage}  \vspace{0.5mm}
    
    \begin{minipage}[t]{0.48\textwidth}
    \centering
    \includegraphics[width=6cm]{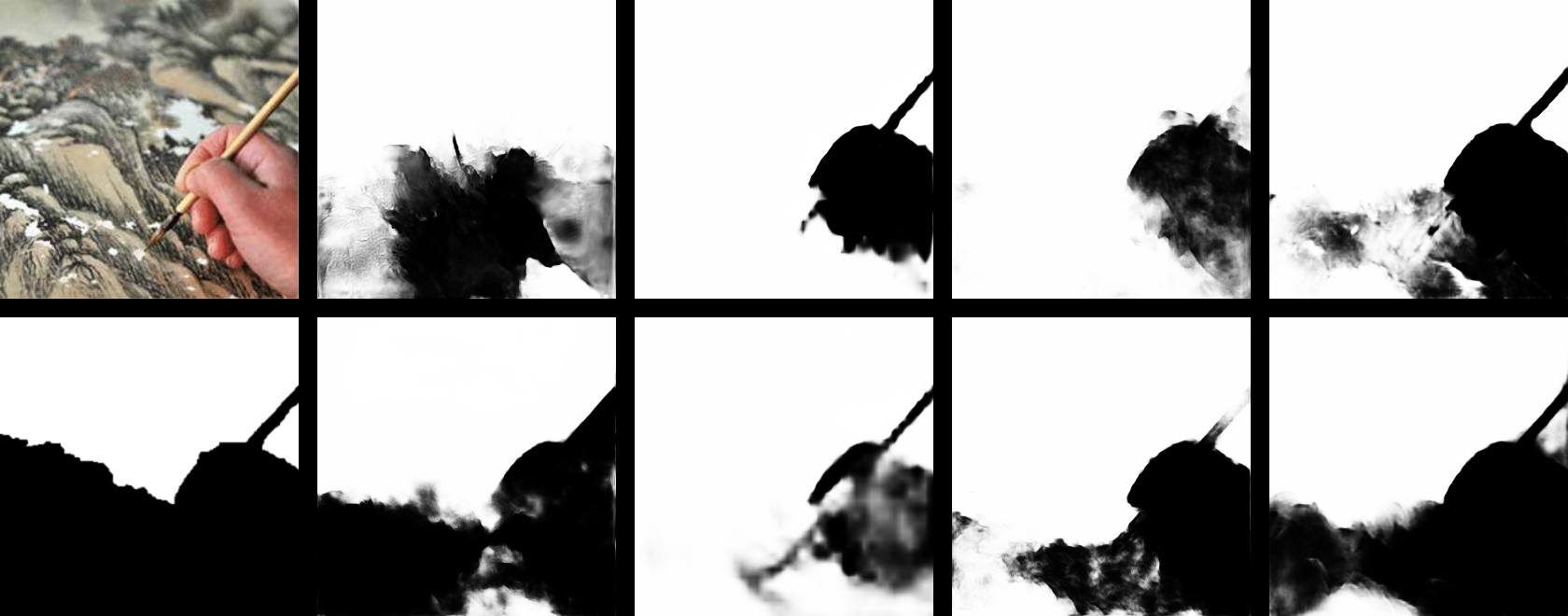}
    \end{minipage}\hspace{-15mm}
    \begin{minipage}[t]{0.48\textwidth}
    \centering
    \includegraphics[width=6cm]{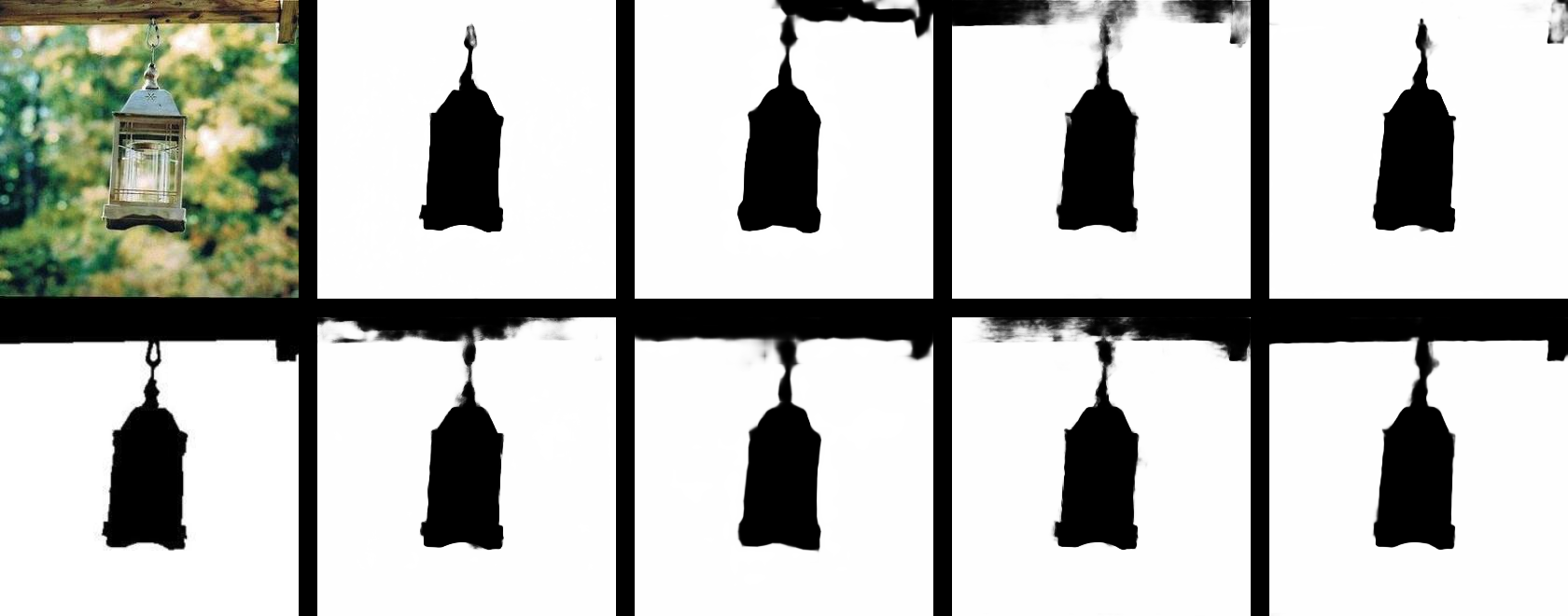}
    \end{minipage}    \vspace{0.5mm}

    \begin{minipage}[t]{0.48\textwidth}
    \centering
    \includegraphics[width=6cm]{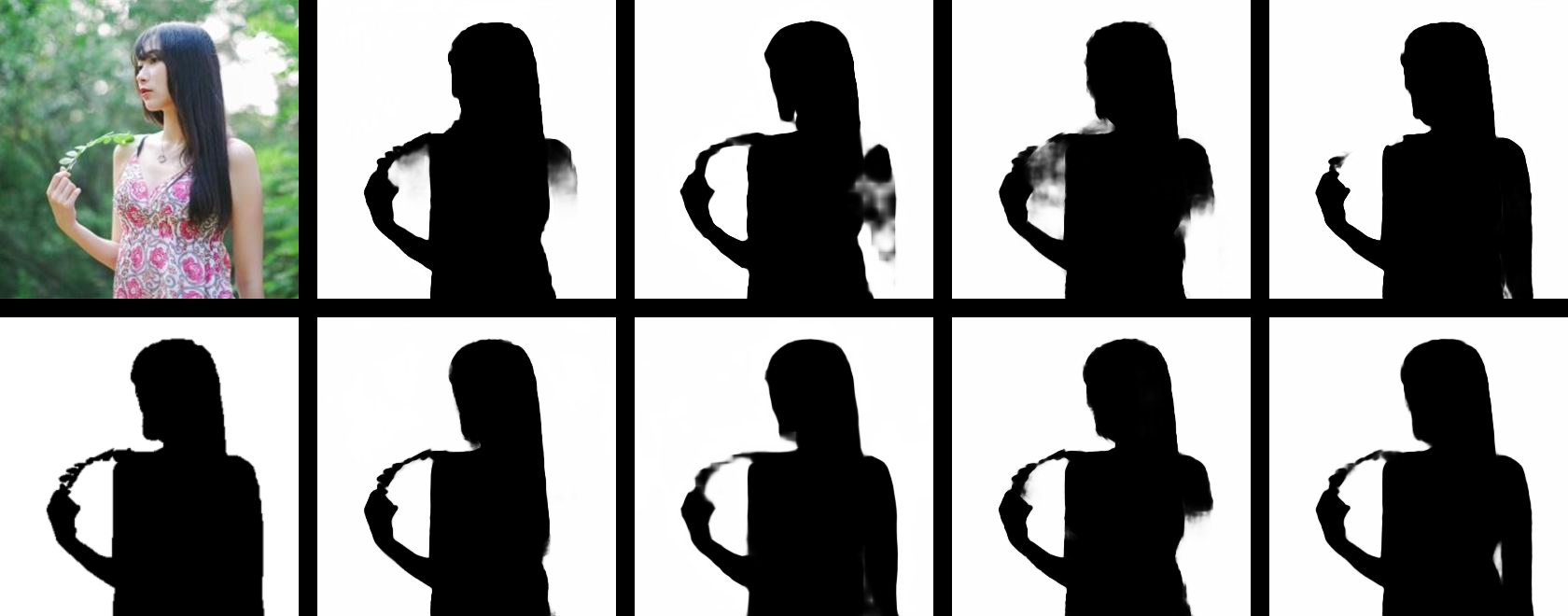}
    \end{minipage}\hspace{-15mm}
    \begin{minipage}[t]{0.48\textwidth}
    \centering
    \includegraphics[width=6cm]{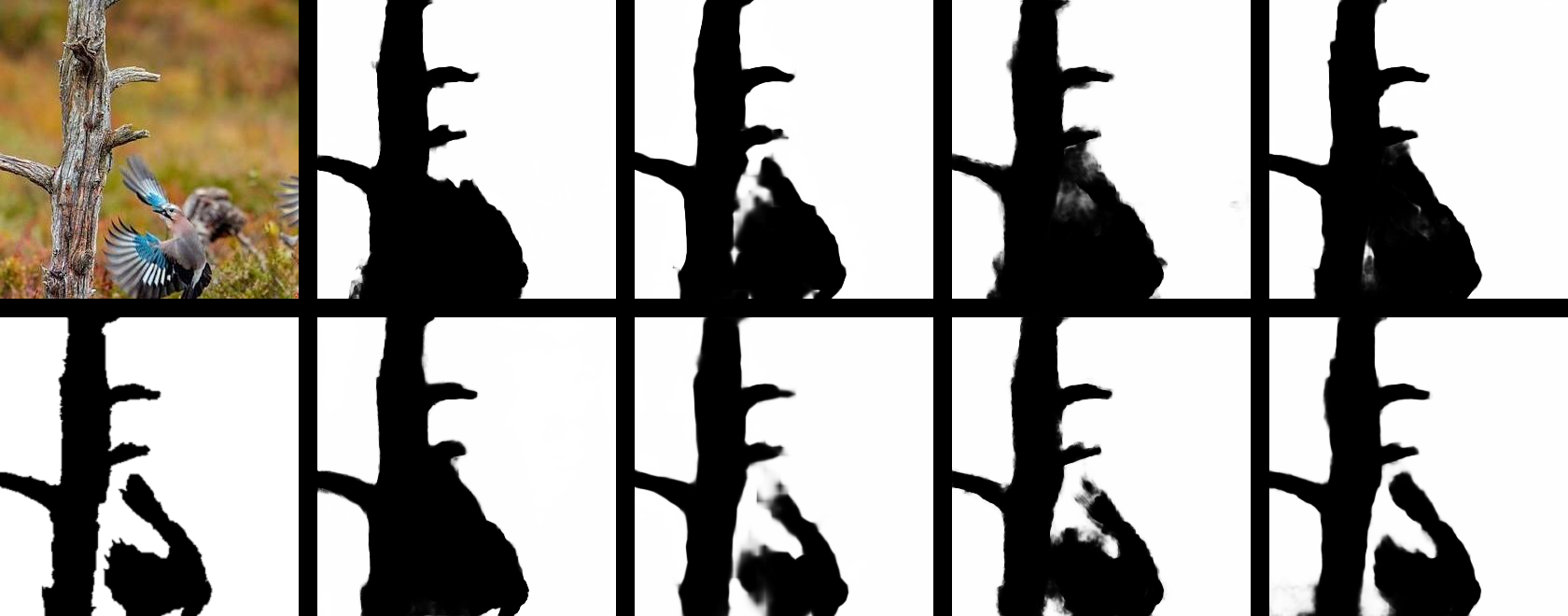}
    \end{minipage}

	\centering
	\caption{Visual comparison of various models results with and w/o our training strategy. There are 12 sets of results, each set contains ten images in 2 rows and 5 columns. In a set, the left row is image and GT, and the right 4 rows are the results from BAS, CPD, MIN, and ours, from left to right. The top one means the results without our training strategy and the bottom one means the results with our training strategy. For instance, the picture in the lower right corner represents our model's results with our training strategy.}
	\label{fig:Quantitative}
\end{figure*}

\subsection{Evaluation Metrics}
In order for a comprehensive evaluation, we choose two common metrics to evaluate our model, $F_{\beta}$\cite{Fmeasure} and MAE\cite{MAE}.
\par $F_{max}$: $F_{\beta}$ is the weighted harmonic mean of precision and recall. The higher the beta is, the more important the recall is considered. We set ${\beta}^2$ to 0.3 as most state-of-the-art methods did. The maximal $F_{\beta}$ values from the PR curve is denoted as $F_{max}$. $F_{\beta}$ is defined as
\begin{equation}
    F_{\beta}=\frac{(1+{\beta}^2)\times {precision}\times {recall}}{{\beta}^2\times {precision}+recall}
\end{equation}

\par MAE: MAE stands for the Mean Absolute Error, it calculates the mean of the absolute value of the difference between the ground truth salient map and the predicted salient map. It is defined as
\begin{equation}
MAE=\frac{1}{H\times{W}}\sum_{1}^{H}\sum_{1}^{W}|P_{(r,c)}-G_{(r,c)}|
\end{equation}
where (r,c) indicates the coordinates in the salient maps, H,W denote the height and width of salient maps, and G, P denote ground truth and prediction salient maps, respectively.

\begin{table*}[ht]
\centering
\renewcommand\tabcolsep{3.0pt}
\caption{Quantitative comparison of F-measure and MAE scores on DBD dataset.}
\label{table:DBDresult}

\begin{tabular}{llcccccccc} 
\hline
Datasets & Metric & SS & LBP & KSFV & DHDE & HiFST & BTBNet & DefusionNET & R2MRF \\ \hline
\multirow{2}{*}{\begin{tabular}[c]{@{}l@{}}Shi et al.'s \\ dataset\end{tabular}} & $maxF_\beta$ & 0.787 & 0.866 & 033733 & 0.850 & 0.856 & 0.892 & 0.917 & 0.927 \\
 & MAE & 0.298 & 0.186 & 0.380 & 0.390 & 0.232 & 0.105 & 0.116 & 0.119 \\ \hline
\multirow{2}{*}{DUT} & $maxF_\beta$ & 0.784 & 0.874 & 0.951 & 0.823 & 0.892 & 0.887 & 0.922 & 0.950 \\
 & MAE & 0.296 & 0.173 & 0.399 & 0.408 & 0.105 & 0.190 & 0.115 & 0.088 \\ \hline
Datasets & Metric & SUN & U2Net & BASnet & MINnetres & SRM-res & CPD\_RES & \begin{tabular}[c]{@{}c@{}}Ours w/o\\ finetune\end{tabular} & Ours \\ \hline
\multirow{2}{*}{\begin{tabular}[c]{@{}l@{}}Shi et al.'s \\ dataset\end{tabular}} & $maxF_\beta$ & \textbf{0.9508} & 0.9435 & 0.9475 & 0.9493 & 0.9336 & 0.9445 & 0.9289 & 0.9365 \\
 & MAE & \textbf{0.0858} & 0.103 & 0.1095 & 0.0935 & 0.1184 & 0.0993 & 0.1134 & 0.1074 \\ \hline
\multirow{2}{*}{DUT} & $maxF_\beta$ & 0.9424 & 0.936 & 0.9353 & 0.9363 & 0.9449 & 0.9257 & 0.9483 & \textbf{0.9538} \\
 & MAE & 0.0801 & 0.0902 & 0.1013 & 0.0964 & 0.0994 & 0.0978 & 0.0840 & \textbf{0.0719} \\ \hline
\end{tabular}
\end{table*}

\subsection{Effectiveness of the transfer learning}
\subsubsection{Quantitative Comparison}
\par To verify the effectiveness of the fine-tuning strategy for DBD task, we compare results of test various methods with our strategy and results without our strategy.
Thanks to the open-source of the current SOD works, we can directly use their well-trained model as the pre-trained model, which including SUN\cite{li2020stacked}, U$^2$-Net\cite{qin2020u2}, u$^2$-Netp\cite{qin2020u2}, BASNet\cite{Qin_2019_CVPR}, MINet\_res\cite{MINet-CVPR2020}, MINet\_vgg\cite{MINet-CVPR2020}, SRM\_res\cite{wang2017stagewise}, CPD\_vgg\cite{Wu_2019_CVPR}, CPD\_res\cite{Wu_2019_CVPR}. For fair comparisons, all models are trained repeatedly and kept their training strategies on SOD dataset , the best result is selected for comparisons.
\par From Table.\ref{table:transfer rlearning result} and Figure.\ref{fig:PRFM}, We could find out that our strategy works on most State-of-the-art deep learning based detection methods. In particular, in terms of the MAE, $F_{\beta}$, PR curves on DUT, and  Shi et al.'s dataset, our strategy works well on the overwhelming majority of methods.

\begin{table}[h]
\caption{Comparison of different models with original training strategy and with our fine-tuning strategy on Shi et al's test set and DUT in terms of $F_{\beta}(\uparrow)$,MAE$(\downarrow)$ \textbf{bold} indicate the higher performance between the two strategies.}

\label{table:transfer rlearning result}
\centering
\renewcommand\tabcolsep{3.0pt}
\begin{tabular}{lccccc}
\hline
methods   & strategy & \multicolumn{2}{c}{shi et al.'s dataset} & \multicolumn{2}{c}{DUT} \\
          &          & max$F_{\beta}$     & MAE             & max$F_{\beta}$     & MAE             \\ \hline
SUN       & original     & 0.9403          & \textbf{0.0858} & 0.9424          & 0.0801          \\
          & finetune & \textbf{0.9484} & 0.0934          & \textbf{0.9483} & \textbf{0.0767} \\ \hline
U$^2$-Net     & original     & 0.9435          & 0.1030          & 0.9306          & 0.0902          \\
          & finetune & \textbf{0.9472} & \textbf{0.0908} & \textbf{0.9525} & \textbf{0.0767} \\ \hline
u$^2$-Netp    & original     & \textbf{0.944}  & 0.1001          & 0.9401          & 0.0972          \\
          & finetune & 0.9411          & \textbf{0.0995} & \textbf{0.9449} & \textbf{0.0824} \\ \hline
BASnet    & original     & \textbf{0.9475} & 0.1018          & 0.9353          & 0.1013          \\
          & finetune & 0.9385          & \textbf{0.0979} & \textbf{0.9373} & \textbf{0.0890} \\ \hline
MINet\_vgg & original     & 0.9451          & 0.1095          & 0.9435          & 0.1001          \\
          & finetune & \textbf{0.9456} & \textbf{0.0973} & \textbf{0.9502} & \textbf{0.0827} \\ \hline
MINet\_res & original     & 0.9493          & \textbf{0.0935} & 0.9363          & 0.0964          \\
          & finetune & \textbf{0.9538} & 0.098           & \textbf{0.9474} & \textbf{0.0932} \\ \hline
SRM\_res   & original     & 0.9336          & 0.1184          & 0.9449          & 0.0994          \\
          & finetune & \textbf{0.937}  & \textbf{0.1113} & \textbf{0.9451} & \textbf{0.0959} \\ \hline
CPD\_vgg  & original     & 0.9386          & 0.1217          & 0.935           & 0.1134          \\
          & finetune & \textbf{0.9395} & \textbf{0.1009} & \textbf{0.9457} & \textbf{0.892}  \\ \hline
CPD\_res  & original     & 0.9445          & 0.0993          & 0.9257          & 0.0978          \\
          & finetune & \textbf{0.9504} & \textbf{0.0969} & \textbf{0.9517} & \textbf{0.0845} \\ \hline
Ours      & original     & 0.9289          & 0.1134          & 0.9483          &  0.0840        \\
          & finetune & \textbf{0.9365} & \textbf{0.1074} & \textbf{0.9538} & \textbf{0.0719} \\ \hline
\end{tabular}
\end{table}

\subsubsection{Qualitative Comparison}
\par Figure.\ref{fig:Quantitative} provides visual comparisons of various models results with and w/o our training strategy. We choose 12 images to represent different situations. According to the results in each set in Figure.\ref{fig:Quantitative}, we can find that our training strategy could easily improve the visual results on various models and preserve better boundary details compared to the results without our training strategy. For instance, in the fourth group of photos on the right, due to the low contrast in the images, the result without our training strategy can not distinguish the left ear and the background in most models.

\begin{table*}[!ht]
	\centering
	\renewcommand\tabcolsep{3.5pt}
	\fontsize{6.5}{8}\selectfont
	\caption{ Performance comparison on six datasets: \textcolor{red}{\uline{Red}}, \textcolor{green}{\uuline{Green}}, and \textcolor{blue}{\textbf{Blue}} indicate the best, second best and third best performance. All methods are annotated by its conference and published year.}
	\label{tab:SOD_performance_comparison}
	\begin{tabular}{ccccccccccccc}
	\toprule
	\multirow{2}{*}{Method}&
	\multicolumn{2}{c}{SOD}&\multicolumn{2}{c}{DUTS-TE}&\multicolumn{2}{c}{ECSSD}&\multicolumn{2}{c}{HKU-IS }&\multicolumn{2}{c}{PASCAL-S}&\multicolumn{2}{c}{DUT-OMRON}\cr
	\cmidrule(lr){2-3} \cmidrule(lr){4-5}\cmidrule(lr){6-7}\cmidrule(lr){8-9}\cmidrule(lr){10-11}\cmidrule(lr){12-13}
	&$maxF_\beta$&MAE&$maxF_\beta$&MAE&$maxF_\beta$&MAE&$maxF_\beta$&MAE&$maxF_\beta$&MAE&$maxF_\beta$&MAE\cr
	\midrule
	{\bf MDF$_{cvpr16}$}	&0.746&0.192				&0.729&0.099					&0.832&0.105
	&0.860&0.129				&0.759&0.142				&0.694&0.142\cr
	
	{\bf AMU$_{cvpr17}$}	&-&-				&0.778&0.085					&0.915&0.059
	&0.899&0.050				&0.841&0.098									&0.744&0.098\cr
	
	{\bf MSRNet$_{cvpr17}$}	&-&-				&0.829&0.061					&0.911&0.054
	&0.914&0.040				&0.858&0.081				&0.782&0.073\cr
	
	{\bf UCF$_{17}$}	&0.808&0.148				&0.773&0.112					&0.903&0.069
						&0.888&0.062				&0.814&0.115		&0.730&0.120\cr					
	{\bf NLDF+$_{cvpr17}$}	&0.841&0.125				&0.813&0.065					&0.905&0.063
						&0.902&0.048				&0.822&0.098					&0.753&0.080\cr
	{\bf C2SNet$_{iccv18}$}	&-&-	&0.811&0.062 	&0.911&0.053	&0.898&0.047 &0.857&0.080			&0.759&0.072\cr		
	{\bf DGRL$_{cvpr18}$}	&0.848&0.106	&0.834&0.051 	&0.925&0.042	&0.913&0.037 &0.848&0.074			&0.779&0.063\cr					
	{\bf RAS$_{eccv18}$}	&0.851&0.124	&0.831&0.059 	&0.921&0.056	&0.913&0.045 &0.829&0.101			&0.786&0.062\cr
	{\bf BMPM$_{cvpr18}$}	&0.856&\textcolor{green}{\uuline{0.108}}								&0.852&0.048									&0.928&0.045	
						&0.921&0.039	&0.850&0.074									&0.774&0.064\cr
	{\bf HRS$_{iccv19}$}	&-&-				&0.843&0.051		&0.920&0.054					&0.913&0.042		&0.852&0.090	&0.762&0.066\cr		
	{\bf CapSal$_{cvpr19}$}	&0.773&0.148				&0.823&0.072		&0.874&0.077
						&0.882&0.062				&0.861&0.073		&0.699&0.101\cr
	{\bf MLMS$_{cvpr19}$}	&0.856&\textcolor{green}{\uuline{0.108}} 				
						&0.852&0.048  		&0.928&0.045  	
	&0.921&0.039  		&0.855&0.074	&\textcolor{red}{\uline{0.852}}&0.048\cr
	
	{\bf MSWS$_{cvpr19}$}	&0.800&0.167			&0.852&0.048		&0.878&0.096	
						&0.856&0.084			&0.786&0.133		&0.767&0.908\cr
	{\bf CPD$_{cvpr19}$}	&\textcolor{blue}{\textbf{0.860}}&0.112		&0.865
	&\textcolor{blue}{\textbf{0.043}}		
	&0.939&0.037	
	&0.925&0.034	
	&\textcolor{blue}{\textbf{0.861}}
	&\textcolor{green}{\uuline{0.071}}	
	&0.797&\textcolor{green}{\uuline{0.056}}\cr
	
	{\bf U$^2$-Net$_{cvpr20}$}&\textcolor{green}{\uuline{0.861}}&\textcolor{green}{\uuline{0.108}}	
	&\textcolor{blue}{\textbf{0.873}}&0.044 		
	&\textcolor{red}{0.951}&\textcolor{red}{\uline{0.033}}
	&\textcolor{red}{\uline{0.935}}&\textcolor{blue}{0.031 } 	
	&0.859&0.074  	
	&\textcolor{green}{\uuline{0.823}}&\textcolor{red}{\uline{0.054}}\cr
	
	{\bf MINet-VGG$_{cvpr20}$}&-&-				&\textcolor{red}{\uline{0.877}}&\textcolor{red}{\uline{0.039}}		
	&\textcolor{blue}{\textbf{0.943}}&\textcolor{blue}{\textbf{0.036}}	
	&\textcolor{blue}{\textbf{0.932}}&\textcolor{red}{\uline{0.030}}	
	&\textcolor{red}{\uline{0.882}}&\textcolor{red}{\uline{0.065}}	
	&0.794&\textcolor{blue}{\textbf{0.057}}\cr
	
	{\bf Ours}			&\textcolor{red}{\textbf{\uline{0.8711}}}&\textcolor{red}{\uline{\bf 0.0828}}
	&\textcolor{green}{\uuline{\bf 0.8711}}&{\textcolor{green}{\uuline{\bf 0.0416}}}
	&\textcolor{green}{\uuline{\bf 0.9487}}&\textcolor{green}{\uuline{\bf 0.0353}}
	&\textcolor{red}{\uline{\bf 0.9355}}&\textcolor{green}{\uuline{\bf 0.0305}}  
	&\textcolor{green}{\uuline{\bf 0.8773}}&\textcolor{blue}{\textbf{\bf 0.0670}}
	&\textcolor{blue}{\textbf{\bf 0.8089}}&{\bf 0.0614}\cr
	
	\bottomrule
	\end{tabular}
\end{table*}
\subsubsection{Comparison of Rate of Convergence}
\par
\begin{figure}[!h]
\begin{center}
\includegraphics[width=0.9\linewidth]{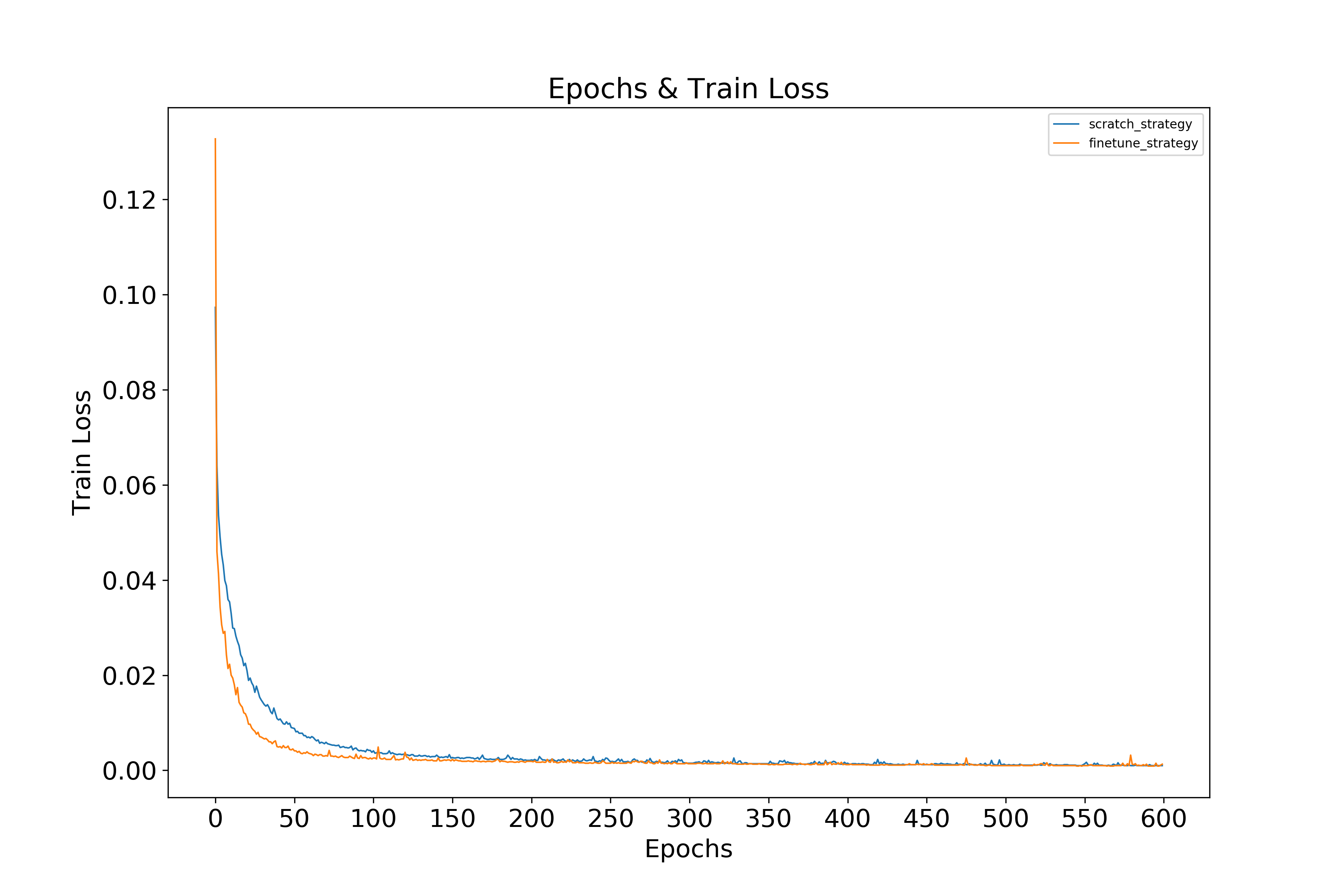}
\end{center}
   \caption{The comparison of training loss of our model with scratch training strategy and with fine-tuning strategy.}
\label{fig:loss}
\end{figure}

 Figure.\ref{fig:loss} shows the train loss curves of the DBD training period with and w/o fine-tune strategy on our model. As shown in Figure.\ref{fig:loss}, with fine-tune strategy, the optimization process time can ease to no more than 200 epochs. At the same time, it will be over 400 epochs without our strategy. Meanwhile, the training error also be reduced. In conclusion, our training strategy could improve the capability of the model in convergence rate and accuracy.

\subsection{Comparison with the state-of-the-art methods on SOD\&DBD}
In the next two sections, we will show the quantitative comparison of MAE and F-measure scores. The visual comparisons and other scores will be shown in the supplementary material.

\subsubsection{Results on DBD task}
In Table.\ref{table:DBDresult}, We compare our method against 14 current models. The table contains two columns of methods, the above methods are the model proposed for DBD task which including SS\cite{tang2016lspectral}, LBP\cite{yi2016lbp}, KSFV\cite{pang2015classifying}, DHDE\cite{park2017unified}, HiFST\cite{golestaneh2017spatially}, BTBNet\cite{zhao2018defocus}, DefusionNet\cite{tang2019defusionnet}, R$^2$MRF\cite{tang2020r2mrf}. The methods below are designed for SOD task which including SUN\cite{li2020stacked}, U$^2$net\cite{qin2020u2}, BASnet\cite{Qin_2019_CVPR}, MINnet\cite{MINet-CVPR2020}, SRM\cite{wang2017stagewise}, CPD\cite{Wu2019CascadedPD}, and we re-train them on DBD training set. From the Table.\ref{table:DBDresult} we could find that SUN method outperforms on the Shi et al.'s dataset and our methods achieves highest result on DUT dataset.


\subsubsection{Results on SOD task}

In order to make a deep comparison, we show our measurement of F-score and MAE in Table \ref{tab:SOD_performance_comparison}, meanwhile comparing our model with 16 current state-of-the-art methods including MDF \cite{Li2016VisualSD}, AMU \cite{Zhang2017AmuletAM}, MSRNet \cite{Li2017InstanceLevelSO}, UCF \cite{Li2017InstanceLevelSO}, NLDF \cite{Luo2017NonlocalDF},C2SNET \cite{Li2018ContourKT},DGRL \cite{Wang2018DetectGR},RAS \cite{Chen2018ReverseAF},BMPM,HRS \cite{Zeng2019TowardsHS},CapSal \cite{Zhang2019CapSalLC}, MLMS \cite{Wu2019AML}, MSWS \cite{zeng2019multi}, CPD \cite{Wu2019CascadedPD}, U$^2$Net \cite{qin2020u2} and MINet-VGG. All results used are obtained from relevant papers or official sites. Our model achieves the best performance on SOD dataset on both metrics and achieves the highest F-score (higher is better) on HKU-IS dataset. Except for the MAE metric on ECSSD dataset and DUT-OMORON dataset, our model achieves either second best or third best results on the remaining datasets.

\section{Conclusions}
\par In this paper, we propse a novel fine-tune strategy to improve the performance of a model on the DBD task. Firstly, we train a model on the SOD training set, then the well-trained model is fed to the DBD training set for finetuning. The experiment shows that our training strategy could outperform the model trained from scratch or a pre-trained backbone. By this strategy, we could utilize the prior of the SOD model to improve the model capability. Besides, we proposed a novel model, it shows robust performance in SOD\&DBD tasks with scale variance. It achieves remarkable results on 2 DBD public tasks and 6 SOD public tasks. 
\par Although our strategy can achieve remarkable promotion to the results of the DBD task on the various models, we can not ignore the fact that the transfer learning method contains various strategies, not just fine-tune strategy. In the future, we could try more transfer methods to explore how to improve the DBD results via SOD prior. 

\clearpage

{\small
\bibliographystyle{ieee_fullname}
\bibliography{egbib}

\begin{thebibliography}{10}\itemsep=-1pt

\bibitem{Fmeasure}
Radhakrishna Achanta, Sheila Hemami, Francisco Estrada, and Sabine
  S{\"u}sstrunk.
\newblock Frequency-tuned salient region detection.
\newblock In {\em The IEEE Conference on Computer Vision and Pattern
  Recognition (CVPR)}, number CONF, pages 1597--1604, 2009.

\bibitem{Achanta2012SLICSC}
R. Achanta, A. Shaji, K. Smith, A. Lucchi, P. Fua, and S. S{\"u}sstrunk.
\newblock Slic superpixels compared to state-of-the-art superpixel methods.
\newblock {\em IEEE Transactions on Pattern Analysis and Machine Intelligence},
  34:2274--2282, 2012.

\bibitem{borji2019salient}
Ali Borji, Ming-Ming Cheng, Qibin Hou, Huaizu Jiang, and Jia Li.
\newblock Salient object detection: A survey.
\newblock {\em Computational visual media}, pages 1--34, 2019.

\bibitem{Chen2018ReverseAF}
S. Chen, Xiuli Tan, B. Wang, and Xuelong Hu.
\newblock Reverse attention for salient object detection.
\newblock {\em ArXiv}, abs/1807.09940, 2018.

\bibitem{davis2008practical}
Harold Davis.
\newblock {\em Practical artistry: light \& exposure for digital
  photographers}.
\newblock " O'Reilly Media, Inc.", 2008.

\bibitem{deng2009imagenet}
Jia Deng, Wei Dong, Richard Socher, Li-Jia Li, Kai Li, and Li Fei-Fei.
\newblock Imagenet: A large-scale hierarchical image database.
\newblock In {\em 2009 IEEE conference on computer vision and pattern
  recognition}, pages 248--255. Ieee, 2009.

\bibitem{ding2020semantic}
Henghui Ding, Xudong Jiang, Bing Shuai, Ai~Qun Liu, and Gang Wang.
\newblock Semantic segmentation with context encoding and multi-path decoding.
\newblock {\em IEEE Transactions on Image Processing}, 29:3520--3533, 2020.

\bibitem{fang2020perceptual}
Yuming Fang, Hanwei Zhu, Yan Zeng, Kede Ma, and Zhou Wang.
\newblock Perceptual quality assessment of smartphone photography.
\newblock In {\em Proceedings of the IEEE/CVF Conference on Computer Vision and
  Pattern Recognition}, pages 3677--3686, 2020.

\bibitem{golestaneh2017spatially}
S~Alireza Golestaneh and Lina~J Karam.
\newblock Spatially-varying blur detection based on multiscale fused and sorted
  transform coefficients of gradient magnitudes.
\newblock In {\em CVPR}, pages 596--605, 2017.

\bibitem{he2019rethinking}
Kaiming He, Ross Girshick, and Piotr Doll{\'a}r.
\newblock Rethinking imagenet pre-training.
\newblock In {\em Proceedings of the IEEE international conference on computer
  vision}, pages 4918--4927, 2019.

\bibitem{He2016DeepRL}
Kaiming He, X. Zhang, Shaoqing Ren, and Jian Sun.
\newblock Deep residual learning for image recognition.
\newblock {\em 2016 IEEE Conference on Computer Vision and Pattern Recognition
  (CVPR)}, pages 770--778, 2016.

\bibitem{Hu2005RobustSA}
Y. Hu, D. Rajan, and L. Chia.
\newblock Robust subspace analysis for detecting visual attention regions in
  images.
\newblock In {\em MULTIMEDIA '05}, 2005.

\bibitem{huang2013cross}
Jui-Ting Huang, Jinyu Li, Dong Yu, Li Deng, and Yifan Gong.
\newblock Cross-language knowledge transfer using multilingual deep neural
  network with shared hidden layers.
\newblock In {\em 2013 IEEE International Conference on Acoustics, Speech and
  Signal Processing}, pages 7304--7308. IEEE, 2013.

\bibitem{kading2016fine}
Christoph K{\"a}ding, Erik Rodner, Alexander Freytag, and Joachim Denzler.
\newblock Fine-tuning deep neural networks in continuous learning scenarios.
\newblock In {\em Asian Conference on Computer Vision}, pages 588--605.
  Springer, 2016.

\bibitem{Lee2016DeepSW}
Gayoung Lee, Yu-Wing Tai, and Junmo Kim.
\newblock Deep saliency with encoded low level distance map and high level
  features.
\newblock {\em 2016 IEEE Conference on Computer Vision and Pattern Recognition
  (CVPR)}, pages 660--668, 2016.

\bibitem{Li2017InstanceLevelSO}
Guanbin Li, Y. Xie, L. Lin, and Y. Yu.
\newblock Instance-level salient object segmentation.
\newblock {\em 2017 IEEE Conference on Computer Vision and Pattern Recognition
  (CVPR)}, pages 247--256, 2017.

\bibitem{Li2015VisualSB}
Guanbin Li and Y. Yu.
\newblock Visual saliency based on multiscale deep features.
\newblock {\em 2015 IEEE Conference on Computer Vision and Pattern Recognition
  (CVPR)}, pages 5455--5463, 2015.

\bibitem{Li2016VisualSD}
Guanbin Li and Y. Yu.
\newblock Visual saliency detection based on multiscale deep cnn features.
\newblock {\em IEEE Transactions on Image Processing}, 25:5012--5024, 2016.

\bibitem{Li2017CNNFS}
Hongyang Li, J. Chen, H. Lu, and Zhizhen Chi.
\newblock Cnn for saliency detection with low-level feature integration.
\newblock {\em Neurocomputing}, 226:212--220, 2017.

\bibitem{li2020stacked}
Junxia Li, Zefeng Pan, Qingshan Liu, and Ziyang Wang.
\newblock Stacked u-shape network with channel-wise attention for salient
  object detection.
\newblock {\em IEEE Transactions on Multimedia}, 2020.

\bibitem{Li2018ABD}
Jia Li, Changqun Xia, and Xiaowu Chen.
\newblock A benchmark dataset and saliency-guided stacked autoencoders for
  video-based salient object detection.
\newblock {\em IEEE transactions on image processing : a publication of the
  IEEE Signal Processing Society}, 27 1:349--364, 2018.

\bibitem{Li2018ContourKT}
Xin Li, Fan Yang, H. Cheng, W. Liu, and Dinggang Shen.
\newblock Contour knowledge transfer for salient object detection.
\newblock In {\em ECCV}, 2018.

\bibitem{Li2014TheSO}
Y. Li, Xiaodi Hou, C. Koch, James~M. Rehg, and A. Yuille.
\newblock The secrets of salient object segmentation.
\newblock {\em 2014 IEEE Conference on Computer Vision and Pattern
  Recognition}, pages 280--287, 2014.

\bibitem{li2020efficient}
Yuelong Li, Mohammad Tofighi, Junyi Geng, Vishal Monga, and Yonina~C Eldar.
\newblock Efficient and interpretable deep blind image deblurring via algorithm
  unrolling.
\newblock {\em IEEE Transactions on Computational Imaging}, 6:666--681, 2020.

\bibitem{long2015learning}
Mingsheng Long, Yue Cao, Jianmin Wang, and Michael Jordan.
\newblock Learning transferable features with deep adaptation networks.
\newblock In {\em International conference on machine learning}, pages 97--105.
  PMLR, 2015.

\bibitem{Luo2017NonlocalDF}
Zhiming Luo, A. Mishra, A. Achkar, Justin~A. Eichel, Shaozi Li, and Pierre-Marc
  Jodoin.
\newblock Non-local deep features for salient object detection.
\newblock {\em 2017 IEEE Conference on Computer Vision and Pattern Recognition
  (CVPR)}, pages 6593--6601, 2017.

\bibitem{oquab2014learning}
Maxime Oquab, Leon Bottou, Ivan Laptev, and Josef Sivic.
\newblock Learning and transferring mid-level image representations using
  convolutional neural networks.
\newblock In {\em Proceedings of the IEEE conference on computer vision and
  pattern recognition}, pages 1717--1724, 2014.

\bibitem{MINet-CVPR2020}
Youwei Pang, Xiaoqi Zhao, Lihe Zhang, and Huchuan Lu.
\newblock Multi-scale interactive network for salient object detection.
\newblock In {\em Proceedings of the IEEE International Conference on Computer
  Vision}, June 2020.

\bibitem{Pang2020MultiScaleIN}
Y. Pang, Xiao-Qi Zhao, L. Zhang, and Huchuan Lu.
\newblock Multi-scale interactive network for salient object detection.
\newblock {\em 2020 IEEE/CVF Conference on Computer Vision and Pattern
  Recognition (CVPR)}, pages 9410--9419, 2020.

\bibitem{pang2015classifying}
Yanwei Pang, Hailong Zhu, Xinyu Li, and Xuelong Li.
\newblock Classifying discriminative features for blur detection.
\newblock {\em IEEE Transactions on Cybernetics}, 46(10):2220--2227, 2015.

\bibitem{park2017unified}
Jinsun Park, Yu-Wing Tai, Donghyeon Cho, and In So~Kweon.
\newblock A unified approach of multi-scale deep and hand-crafted features for
  defocus estimation.
\newblock In {\em Proceedings of the IEEE Conference on Computer Vision and
  Pattern Recognition}, pages 1736--1745, 2017.

\bibitem{MAE}
Federico Perazzi, Philipp Kr{\"a}henb{\"u}hl, Yael Pritch, and Alexander
  Hornung.
\newblock Saliency filters: Contrast based filtering for salient region
  detection.
\newblock In {\em The IEEE Conference on Computer Vision and Pattern
  Recognition (CVPR)}, pages 733--740, 2012.

\bibitem{purohit2018learning}
Kuldeep Purohit, Anshul~B Shah, and AN Rajagopalan.
\newblock Learning based single image blur detection and segmentation.
\newblock In {\em 2018 25th IEEE International Conference on Image Processing
  (ICIP)}, pages 2202--2206. IEEE, 2018.

\bibitem{qian2020bggan}
Ming Qian, Congyu Qiao, Jiamin Lin, Zhenyu Guo, Chenghua Li, Cong Leng, and
  Jian Cheng.
\newblock Bggan: Bokeh-glass generative adversarial network for rendering
  realistic bokeh.
\newblock {\em arXiv preprint arXiv:2011.02242}, 2020.

\bibitem{qin2020u2}
Xuebin Qin, Zichen Zhang, Chenyang Huang, Masood Dehghan, Osmar~R Zaiane, and
  Martin Jagersand.
\newblock U2-net: Going deeper with nested u-structure for salient object
  detection.
\newblock {\em Pattern Recognition}, 106:107404, 2020.

\bibitem{Qin_2019_CVPR}
Xuebin Qin, Zichen Zhang, Chenyang Huang, Chao Gao, Masood Dehghan, and Martin
  Jagersand.
\newblock Basnet: Boundary-aware salient object detection.
\newblock In {\em The IEEE Conference on Computer Vision and Pattern
  Recognition (CVPR)}, June 2019.

\bibitem{redmon2018yolov3}
Joseph Redmon and Ali Farhadi.
\newblock Yolov3: An incremental improvement.
\newblock {\em arXiv preprint arXiv:1804.02767}, 2018.

\bibitem{shi2014discriminative}
Jianping Shi, Li Xu, and Jiaya Jia.
\newblock Discriminative blur detection features.
\newblock In {\em Proceedings of the IEEE Conference on Computer Vision and
  Pattern Recognition}, pages 2965--2972, 2014.

\bibitem{shi2015just}
Jianping Shi, Li Xu, and Jiaya Jia.
\newblock Just noticeable defocus blur detection and estimation.
\newblock In {\em Proceedings of the IEEE Conference on Computer Vision and
  Pattern Recognition}, pages 657--665, 2015.

\bibitem{Shi2016HierarchicalIS}
J. Shi, Q. Yan, L. Xu, and J. Jia.
\newblock Hierarchical image saliency detection on extended cssd.
\newblock {\em IEEE Transactions on Pattern Analysis and Machine Intelligence},
  38:717--729, 2016.

\bibitem{su2011blurred}
Bolan Su, Shijian Lu, and Chew~Lim Tan.
\newblock Blurred image region detection and classification.
\newblock In {\em Proceedings of the 19th ACM international conference on
  Multimedia}, pages 1397--1400, 2011.

\bibitem{tan2018survey}
Chuanqi Tan, Fuchun Sun, Tao Kong, Wenchang Zhang, Chao Yang, and Chunfang Liu.
\newblock A survey on deep transfer learning.
\newblock In {\em International conference on artificial neural networks},
  pages 270--279. Springer, 2018.

\bibitem{tang2020r2mrf}
Chang Tang, Xinwang Liu, Xinzhong Zhu, En Zhu, Kun Sun, Pichao Wang, Lizhe
  Wang, and Albert~Y Zomaya.
\newblock R$^2$mrf: Defocus blur detection via recurrently refining multi-scale
  residual features.
\newblock In {\em AAAI}, pages 12063--12070, 2020.

\bibitem{tang2016lspectral}
Chang Tang, Jin Wu, Yonghong Hou, Pichao Wang, and Wanqing Li.
\newblock A spectral and spatial approach of coarse-to-fine blurred image
  region detection.
\newblock {\em IEEE Signal Processing Letters}, 23(11):1652--1656, 2016.

\bibitem{tang2019defusionnet}
Chang Tang, Xinzhong Zhu, Xinwang Liu, Lizhe Wang, and Albert Zomaya.
\newblock Defusionnet: Defocus blur detection via recurrently fusing and
  refining multi-scale deep features.
\newblock In {\em Proceedings of the IEEE Conference on Computer Vision and
  Pattern Recognition}, pages 2700--2709, 2019.

\bibitem{tzeng2017adversarial}
Eric Tzeng, Judy Hoffman, Kate Saenko, and Trevor Darrell.
\newblock Adversarial discriminative domain adaptation.
\newblock In {\em Proceedings of the IEEE conference on computer vision and
  pattern recognition}, pages 7167--7176, 2017.

\bibitem{Wang2015DeepNF}
L. Wang, H. Lu, X. Ruan, and Ming-Hsuan Yang.
\newblock Deep networks for saliency detection via local estimation and global
  search.
\newblock {\em 2015 IEEE Conference on Computer Vision and Pattern Recognition
  (CVPR)}, pages 3183--3192, 2015.

\bibitem{wang2017learning}
Lijun Wang, Huchuan Lu, Yifan Wang, Mengyang Feng, Dong Wang, Baocai Yin, and
  Xiang Ruan.
\newblock Learning to detect salient objects with image-level supervision.
\newblock In {\em Proceedings of the IEEE Conference on Computer Vision and
  Pattern Recognition}, pages 136--145, 2017.

\bibitem{wang2017stagewise}
Tiantian Wang, Ali Borji, Lihe Zhang, Pingping Zhang, and Huchuan Lu.
\newblock A stagewise refinement model for detecting salient objects in images.
\newblock In {\em Proceedings of the IEEE International Conference on Computer
  Vision}, pages 4019--4028, 2017.

\bibitem{Wang2018DetectGR}
Tiantian Wang, L. Zhang, S. Wang, H. Lu, G. Yang, X. Ruan, and A. Borji.
\newblock Detect globally, refine locally: A novel approach to saliency
  detection.
\newblock {\em 2018 IEEE/CVF Conference on Computer Vision and Pattern
  Recognition}, pages 3127--3135, 2018.

\bibitem{Wu2019AML}
R. Wu, Mengyang Feng, Wenlong Guan, D. Wang, H. Lu, and E. Ding.
\newblock A mutual learning method for salient object detection with
  intertwined multi-supervision.
\newblock {\em 2019 IEEE/CVF Conference on Computer Vision and Pattern
  Recognition (CVPR)}, pages 8142--8151, 2019.

\bibitem{Wu_2019_CVPR}
Zhe Wu, Li Su, and Qingming Huang.
\newblock Cascaded partial decoder for fast and accurate salient object
  detection.
\newblock In {\em The IEEE Conference on Computer Vision and Pattern
  Recognition (CVPR)}, June 2019.

\bibitem{Wu2019CascadedPD}
Zhe Wu, L. Su, and Q. Huang.
\newblock Cascaded partial decoder for fast and accurate salient object
  detection.
\newblock {\em 2019 IEEE/CVF Conference on Computer Vision and Pattern
  Recognition (CVPR)}, pages 3902--3911, 2019.

\bibitem{xu2017estimating}
Guodong Xu, Yuhui Quan, and Hui Ji.
\newblock Estimating defocus blur via rank of local patches.
\newblock In {\em Proceedings of the IEEE International Conference on Computer
  Vision}, pages 5371--5379, 2017.

\bibitem{xu2017unified}
Yonghui Xu, Sinno~Jialin Pan, Hui Xiong, Qingyao Wu, Ronghua Luo, Huaqing Min,
  and Hengjie Song.
\newblock A unified framework for metric transfer learning.
\newblock {\em IEEE Transactions on Knowledge and Data Engineering},
  29(6):1158--1171, 2017.

\bibitem{yan2020deep}
Chenggang Yan, Biao Gong, Yuxuan Wei, and Yue Gao.
\newblock Deep multi-view enhancement hashing for image retrieval.
\newblock {\em IEEE Transactions on Pattern Analysis and Machine Intelligence},
  2020.

\bibitem{yang2013saliency}
Chuan Yang, Lihe Zhang, Huchuan Lu, Xiang Ruan, and Ming-Hsuan Yang.
\newblock Saliency detection via graph-based manifold ranking.
\newblock In {\em Proceedings of the IEEE conference on computer vision and
  pattern recognition}, pages 3166--3173, 2013.

\bibitem{yi2016lbp}
Xin Yi and Mark Eramian.
\newblock Lbp-based segmentation of defocus blur.
\newblock {\em IEEE transactions on image processing}, 25(4):1626--1638, 2016.

\bibitem{Zeng2019TowardsHS}
Y. Zeng, Pingping Zhang, Jianming Zhang, Zhe Lin, and H. Lu.
\newblock Towards high-resolution salient object detection.
\newblock {\em 2019 IEEE/CVF International Conference on Computer Vision
  (ICCV)}, pages 7233--7242, 2019.

\bibitem{zeng2019multi}
Yu Zeng, Yunzhi Zhuge, Huchuan Lu, Lihe Zhang, Mingyang Qian, and Yizhou Yu.
\newblock Multi-source weak supervision for saliency detection.
\newblock In {\em Proceedings of the IEEE Conference on Computer Vision and
  Pattern Recognition}, pages 6074--6083, 2019.

\bibitem{Zhang2018ABM}
L. Zhang, J. Dai, H. Lu, You He, and G. Wang.
\newblock A bi-directional message passing model for salient object detection.
\newblock {\em 2018 IEEE/CVF Conference on Computer Vision and Pattern
  Recognition}, pages 1741--1750, 2018.

\bibitem{Zhang2019CapSalLC}
Lu Zhang, Jianming Zhang, Zhe Lin, H. Lu, and Y. He.
\newblock Capsal: Leveraging captioning to boost semantics for salient object
  detection.
\newblock {\em 2019 IEEE/CVF Conference on Computer Vision and Pattern
  Recognition (CVPR)}, pages 6017--6026, 2019.

\bibitem{Zhang2017AmuletAM}
Pingping Zhang, D. Wang, H. Lu, Hongyu Wang, and X. Ruan.
\newblock Amulet: Aggregating multi-level convolutional features for salient
  object detection.
\newblock {\em 2017 IEEE International Conference on Computer Vision (ICCV)},
  pages 202--211, 2017.

\bibitem{zhang2018learning}
Shanghang Zhang, Xiaohui Shen, Zhe Lin, Radom{\'\i}r M{\v{e}}ch, Joao~P
  Costeira, and Jos{\'e}~MF Moura.
\newblock Learning to understand image blur.
\newblock In {\em Proceedings of the IEEE Conference on Computer Vision and
  Pattern Recognition}, pages 6586--6595, 2018.

\bibitem{Zhang2018ProgressiveAG}
X. Zhang, Tiantian Wang, J. Qi, H. Lu, and G. Wang.
\newblock Progressive attention guided recurrent network for salient object
  detection.
\newblock {\em 2018 IEEE/CVF Conference on Computer Vision and Pattern
  Recognition}, pages 714--722, 2018.

\bibitem{zhao2018defocus}
Wenda Zhao, Fan Zhao, Dong Wang, and Huchuan Lu.
\newblock Defocus blur detection via multi-stream bottom-top-bottom fully
  convolutional network.
\newblock In {\em Proceedings of the IEEE conference on computer vision and
  pattern recognition}, pages 3080--3088, 2018.

\bibitem{zhao2019enhancing}
Wenda Zhao, Bowen Zheng, Qiuhua Lin, and Huchuan Lu.
\newblock Enhancing diversity of defocus blur detectors via cross-ensemble
  network.
\newblock In {\em Proceedings of the IEEE Conference on Computer Vision and
  Pattern Recognition}, pages 8905--8913, 2019.

\end{thebibliography}
}

\end{document}